\documentclass[sn-mathphys]{sn-jnl}% Math and Physical Sciences Reference Style
\jyear{2021}%

\theoremstyle{thmstyleone}%
\theoremstyle{thmstyletwo}%

\theoremstyle{thmstylethree}%

\usepackage{graphicx}
\usepackage{lineno}
\begin{document}

\title[GateNet: A novel Neural Network Architecture for automated FC Gating]{GateNet: A novel Neural Network Architecture for Automated Flow Cytometry Gating}


\author*[1]{\fnm{Lukas} \sur{Fisch}}\email{l.fisch@wwu.de}
\author[2]{\fnm{Michael O.} \sur{Heming}}
\author[2]{\fnm{Andreas} \sur{Schulte-Mecklenbeck}}
\author[2]{\fnm{Catharina C.} \sur{Gross}}
\author[1]{\fnm{Stefan} \sur{Zumdick}}
\author[1]{\fnm{Carlotta} \sur{Barkhau}}
\author[1]{\fnm{Daniel} \sur{Emden}}
\author[1,3]{\fnm{Jan} \sur{Ernsting}}
\author[1]{\fnm{Ramona} \sur{Leenings}}
\author[1]{\fnm{Kelvin} \sur{Sarink}}
\author[1]{\fnm{Nils R.} \sur{Winter}}
\author[1]{\fnm{Udo} \sur{Dannlowski}}
\author[2]{\fnm{Heinz} \sur{Wiendl}}
\author[2]{\fnm{Gerd} \sur{Meyer zu Hörste}}
\equalcont{These authors contributed equally to this work.}
\author[1,2]{\fnm{Tim} \sur{Hahn}}
\equalcont{These authors contributed equally to this work.}

\affil*[1]{\orgdiv{University of Münster}, \orgname{Institute for Translational Psychiatry}, \orgaddress{\city{Münster}, \country{Germany}}}
\affil[2]{\orgdiv{Department of Neurology with Institute of Translational Neurology}, \orgname{University and University Hospital Münster}, \orgaddress{\city{Münster}, \country{Germany}}}
\affil[3]{\orgdiv{Faculty of Mathematics and Computer Science}, \orgname{University of Münster}, \orgaddress{\city{Münster}, \country{Germany}}}

\abstract{Flow cytometry is widely used to identify cell populations in patient-derived fluids such as peripheral blood (PB) or cerebrospinal fluid (CSF). While ubiquitous in research and clinical practice, flow cytometry requires gating, i.e. cell type identification which requires labor-intensive and error-prone manual adjustments. To facilitate this process, we designed GateNet, the first neural network architecture enabling full end-to-end automated gating without the need to correct for batch effects. We train GateNet with over 8,000,000 events based on N=127 PB and CSF samples which were manually labeled independently by four experts. We show that for novel, unseen samples, GateNet achieves human-level performance (F1 score ranging from 0.910 to 0.997). In addition we apply GateNet to a publicly available dataset confirming generalization with an F1 score of 0.936. As our implementation utilizes graphics processing units (GPU), gating only needs 15 microseconds per event. Importantly, we also show that GateNet only requires $\approx$10 samples to reach human-level performance, rendering it widely applicable in all domains of flow cytometry.}


\maketitle

\section{Introduction}

\begin{figure}[ht]
\centering
\includegraphics[width=\textwidth]{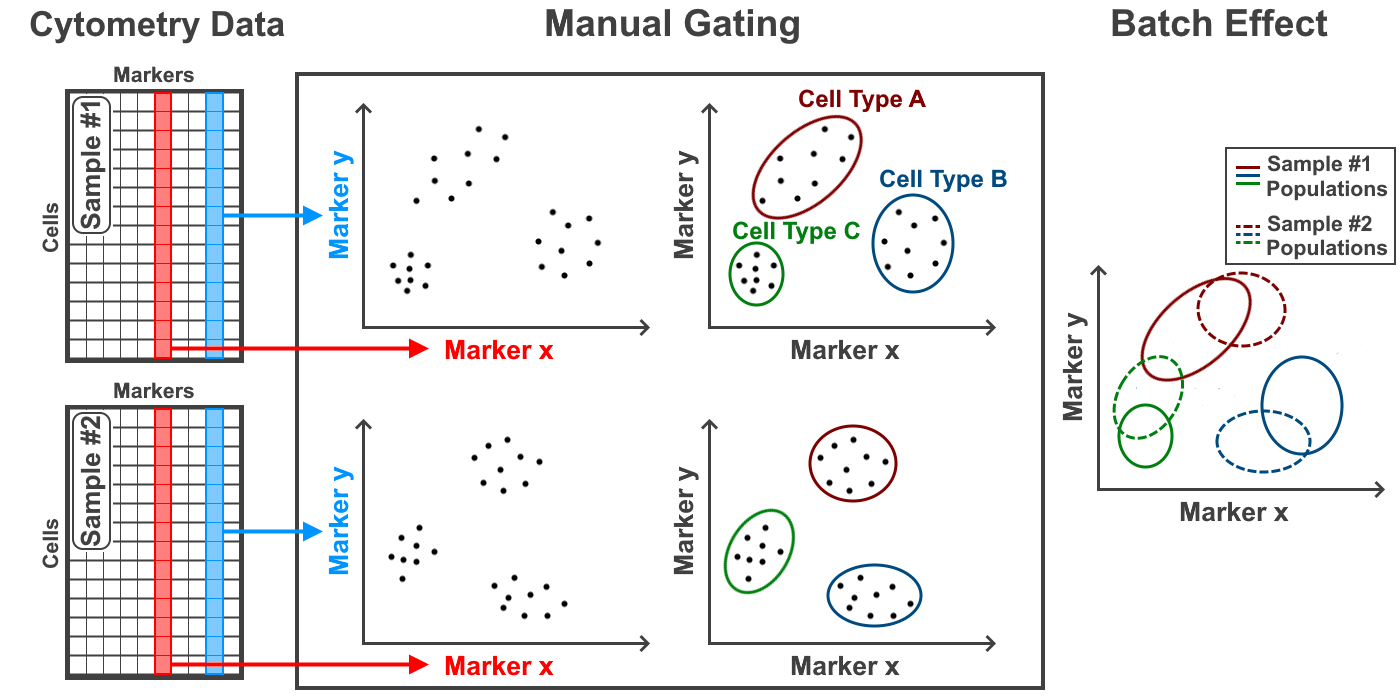}
\caption{Schematic manual gating workflow which corrects for measurement variances across samples caused by the batch effect.} 
\label{batch_effect}
\end{figure}

Flow cytometry (FC) is an analytical technique which is used in biological research to identify cell types and in the clinical context to diagnose human diseases including hematological malignancies\cite{cra08}. FC characterizes cell types by measuring the light scatter and fluorescence emission properties of fluorochrome-labeled antibodies from each of the thousands of cells a sample contains\cite{bau00}. Based on the measured intensity of the fluorescence and the light scatter of these cell events, cells are distinguished from contaminants, and then each cell is classified into a specific cell population. Traditionally, this classification is done by manually identifying and partitioning (i.e. ’gating’) these populations based on visual inspection of mostly two-dimensional intensity histograms of two respective fluorescence emission detectors (Fig. \ref{batch_effect}).

The first obstacle during gating is the batch effect, i.e. technical variance of event measurements across samples, caused e.g. by the variability of the staining procedure or by the decay of the exciting laser and the fluorescence emissions of fluorophore-bound antibodies. Due to this variance, the locations of cell populations in the intensity histograms differ across samples (Fig. \ref{batch_effect}). Therefore, the usage of fixed gates across samples results in low quality partitioning of the respective populations. To reliably capture the population across samples, the gates have to be adjusted manually for each sample which is, both, time consuming and introduces a high degree of subjectivity into the process. This subjectivity impairs reproducibility of results even among experts\cite{gra20,gra21} necessitating centralized manual gating in cross-center studies\cite{fin16}. Most importantly, this issue hampers routine clinical gating at scale, which is relevant for thousands of patients everyday.

The second obstacle pertains to finding all cell populations and subpopulations in a given set of samples. This issue is aggravated by the increasing number of markers which can be measured with modern cytometry technology, such as cytometry by time of flight (CyTOF). To assist in the identification of novel cell populations in samples with $>$ 50 markers measured, a plethora of dimensionality reduction and clustering methods have been proposed\cite{qui07}. These methods are used in conjunction with preprocessing steps which aim to minimize the batch effect and align the position of populations across samples\cite{sch19,sha17} resulting in semi-automated gating.

While helpful for exploration, these semi-automated clustering approaches are prone to instability and therefore still need significant human input. Specifically, these methods are sensitive to random fluctuations and parameter settings such that they do not converge to a consistent clustering\cite{mel17}. In consequence, the choice of methods and respective parameters introduces additional subjectivity into the gating process. Because of these issues the majority of clinical laboratories do not use automation approaches and rely on manual gating despite its time-consuming nature and suboptimal reliability\cite{che21}.

In contrast to the numerous gating methods which employ unsupervised learning, there exist only two works which approach gating with supervised learning of neural networks\cite{che22, li17}. The respective approaches DGCyTOF and DeepCyTOF use fully connected neural networks which rely on a single event measurement to predict the respective gate class of the event.

\begin{figure}[ht]
\centering
\includegraphics[width=\textwidth]{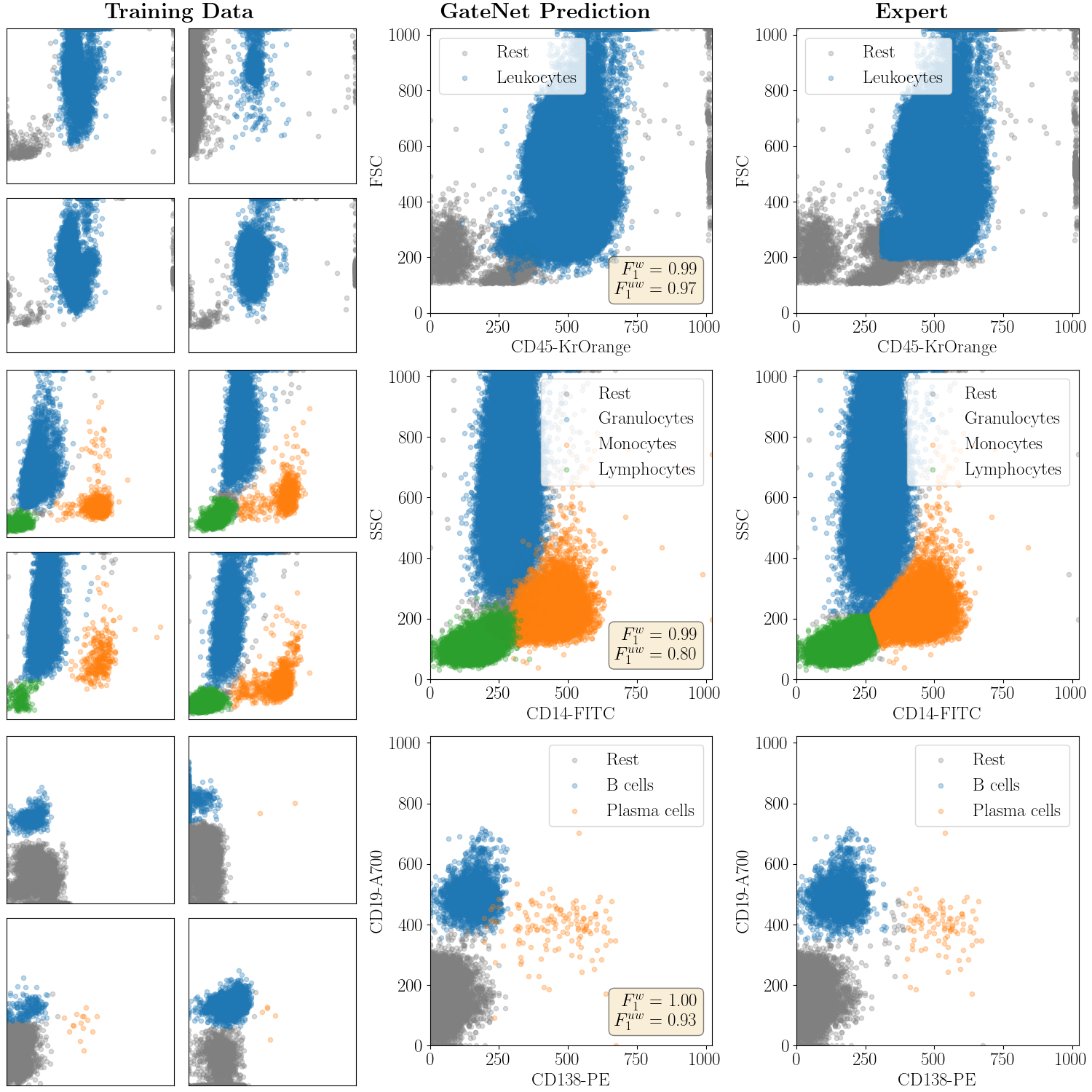}
\caption{GateNet trained on four samples (left) and successively applied to an unseen sample (middle) for three subdatasets of the RheumaFlow dataset. For comparison, the experts manual gating of the respective samples (right) and the weighted and unweighted F1 scores ($F_1^W$ and $F_1^{UW}$) between the prediction and the experts gating are shown.} 
\label{teaser}
\end{figure}

Aiming to fully automate gating, we propose GateNet, the first neural network which was specifically designed for gating. It enabled end-to-end automated gating while automatically correcting for the batch effect using expert gated samples for model training. Its neural network architecture utilizes both the single event measurement as well as the context of events of the sample. This architectural feature allowed it to classify cell populations while reliably correcting for measurement variances of samples (Fig. \ref{teaser}).

In this work, GateNet was applied to a single center dataset with N=127 samples named RheumaFlow, containing hierarchical cell populations, and to the Normal Donors (ND) dataset with N=30 samples, previously published in context of the FlowCap I (Critical Assessment of Population Identification Methods) challenge\cite{flowcap}. Performance was cross-validated using the weighted and unweighted F1 score to evaluate overall as well as rare cell type classification performance\cite{web16}. In addition, the gating reliability of GateNet was compared to the reliability of the four human experts across all samples of the RheumaFlow dataset. To deepen the understanding of GateNet’s process of gating, differences between human and GateNet gating were investigated in detail. Finally, the impact of the number of training samples on the performance was examined.

\section{Results}

Across validation folds, GateNet performance (mean weighted F1 score) ranged from 0.886 up to 0.997 for all datasets (Supplement Table \ref{results}). GateNet achieved weighted F1 scores of 0.910 ($\pm$ 0.034) and 0.945 ($\pm$ 0.023) in the subdatasets CD56CD16 and CD8HLADR of the cerebrospinal fluid (CSF) samples, thus showing high accuracy even for very low number of events (100 and 440 events per sample respectively; see Table \ref{datasets}). In the publicly available Normal donors (ND) dataset, a weighted F1 score of 0.936 ($\pm$ 0.010) was achieved, confirming excellent generalization across datasets.

\begin{figure}[ht]
\centering
\includegraphics[width=\textwidth]{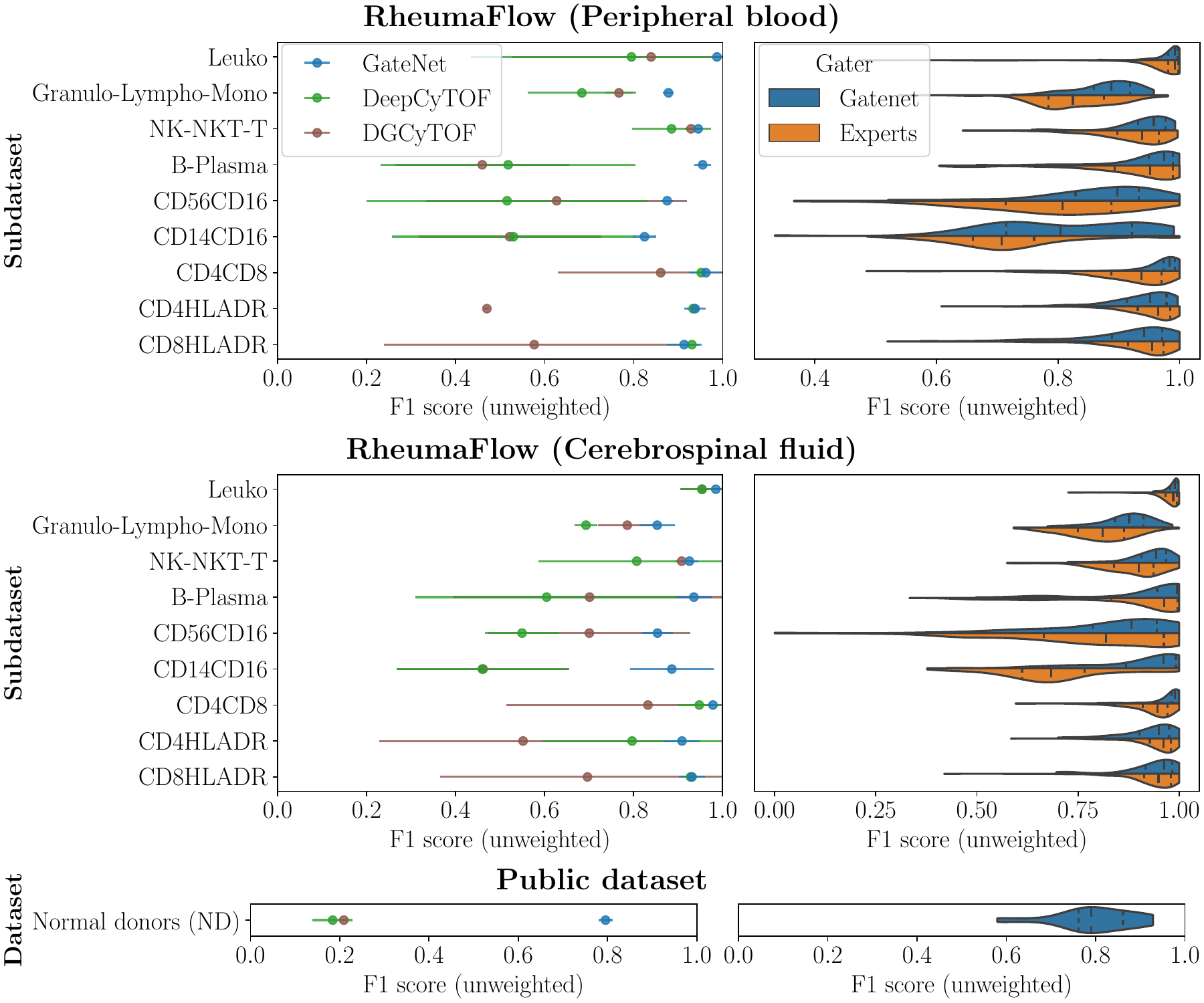}
\caption{Left: Unweighted F1 score of GateNet (blue) obtained during 5-fold cross-validation in the validation sets compared to DeepCyTOF (green) and DGCyTOF (brown). Standard deviations across folds are shown as error bars. Right: GateNets unweighted F1 score distribution (blue) across validation samples (5-fold cross-validation). The respective human expert distribution (orange) is calculated in a leave-one-out-scheme with the ground truth with respect to one expert being the mean of all remaining experts.} 
\label{violins}
\end{figure}

The unweighted F1 scores ranged from 0.796 to 0.987 indicating a robust performance even for datasets with highly imbalanced data (Fig. \ref{violins} and Supplement Table \ref{results}). Even in the B-Plasma subdataset of the peripheral blood (PB) samples which had the highest class imbalance with only 0.023\% of events being Plasma cells, GateNet managed to achieve an unweighted F1 Score of 0.955 ($\pm$ 0.019). On top of that, GateNet outperformed both DGCyTOF and DeepCyTOF (see Section \ref{secA2}) which showed that the specialized network architecture in conjunction with the imbalanced data methods (see Section \ref{ssec:imbalance}) enables the reliable identification of rare cell types.

\subsection{Comparison to human experts}

Manual gatings of the RheumaFlow dataset by four independent human experts were used as ground truth for model validation. The human experts performance metrics were measured using a leave-one-out scheme – i.e. the gating of expert 1 was compared to the mean of the gatings of experts 2, 3 and 4, the gating of expert 2 was compared to the mean of experts 1, 3 and 4 etc. In Figure \ref{violins} (right) the resulting metric distribution of the human experts across all samples was compared to the respective metric distribution of GateNet.

First, the human and the GateNet sample performances were similarly distributed, showing that the performance variance of GateNet mirrored the experts gating disagreement present in the training data. Considering the median of the distributions, GateNet deviated nominally less from the human experts median than the individual human experts in 15 of the 18 subdatasets.

The distributions were long tailed, meaning that there exist few samples for which the experts disagree strongly while most of the gating is consistent across samples. The CD14CD16 subdataset in PB and CSF samples showed the lowest agreement among experts with a median F1 score (unweighted) of 0.707 and 0.684 respectively. This put GateNets comparatively low performance in this gate in perspective and suggested that it gated more reliably than human experts.

\subsection{Learning high-dimensional gating from 2-dimensional training data}

GateNet exploited the measurements of all markers present in the data to form its predictions. Even though it was trained with manual gates which were done purely based on two-dimensional panels, it learned to incorporate the measurements across all markers to achieve an optimal separation of populations across all samples in the training data.

\begin{figure}[ht]
\includegraphics[width=\textwidth]{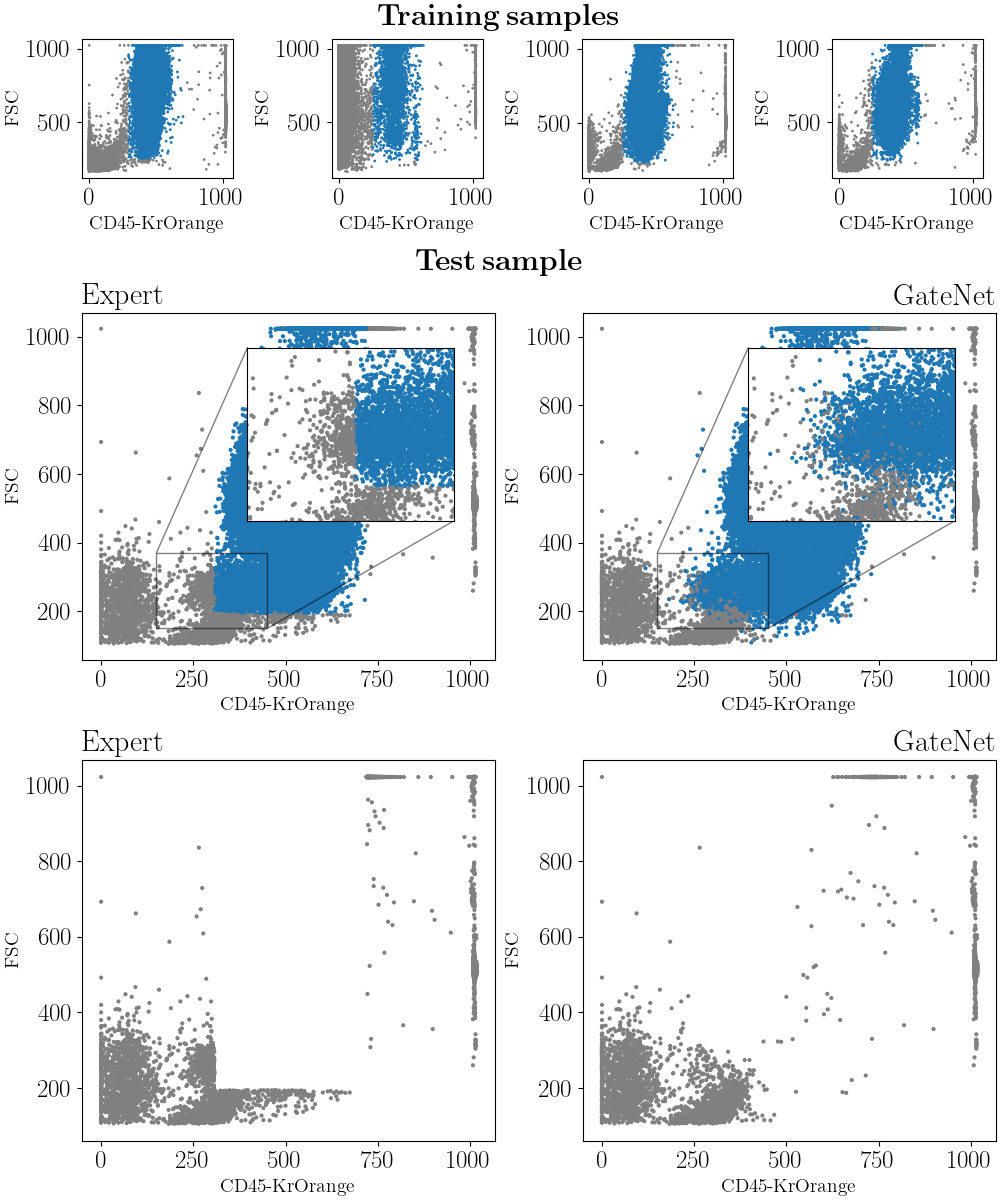}
\caption{GateNet trained on four training samples (top) and subsequently applied to a test sample (bottom right). Expert gating of the test sample is shown in the bottom left.} \label{3d_gate}
\end{figure}

To showcase GateNet’s generalization ability, we selected four training samples which show little population overlap in the two-dimensional panel. Due to this little overlap the respective samples could be gated precisely with traditional manual gating (Fig. \ref{3d_gate}). GateNet is trained using these four precisely gated samples and subsequently applied to a test sample which is hard to gate manually, since populations overlap in the respective panel (Fig. \ref{3d_gate}, bottom left). GateNets gating (Fig. \ref{3d_gate}, bottom right) strongly suggested that it successfully separated these overlapping populations.

As shown in Figure \ref{teaser}, such high precision gating using few training samples can be reached across datasets. Despite the high quality gating, the unweighted F1 scores of the three examples was  not among the best samples in the respective datasets (Fig. \ref{violins}). This suggested that GateNet’s high precision gating challenged the presumed ground truth being the consensus of the four expert gaters. Therefore, higher F1 scores could imply GateNet being overfitted to the human experts gating strategy including its errors.

\subsection{Small training data}
\label{ssec:small_data}

\begin{figure}[ht]
\includegraphics[width=\textwidth]{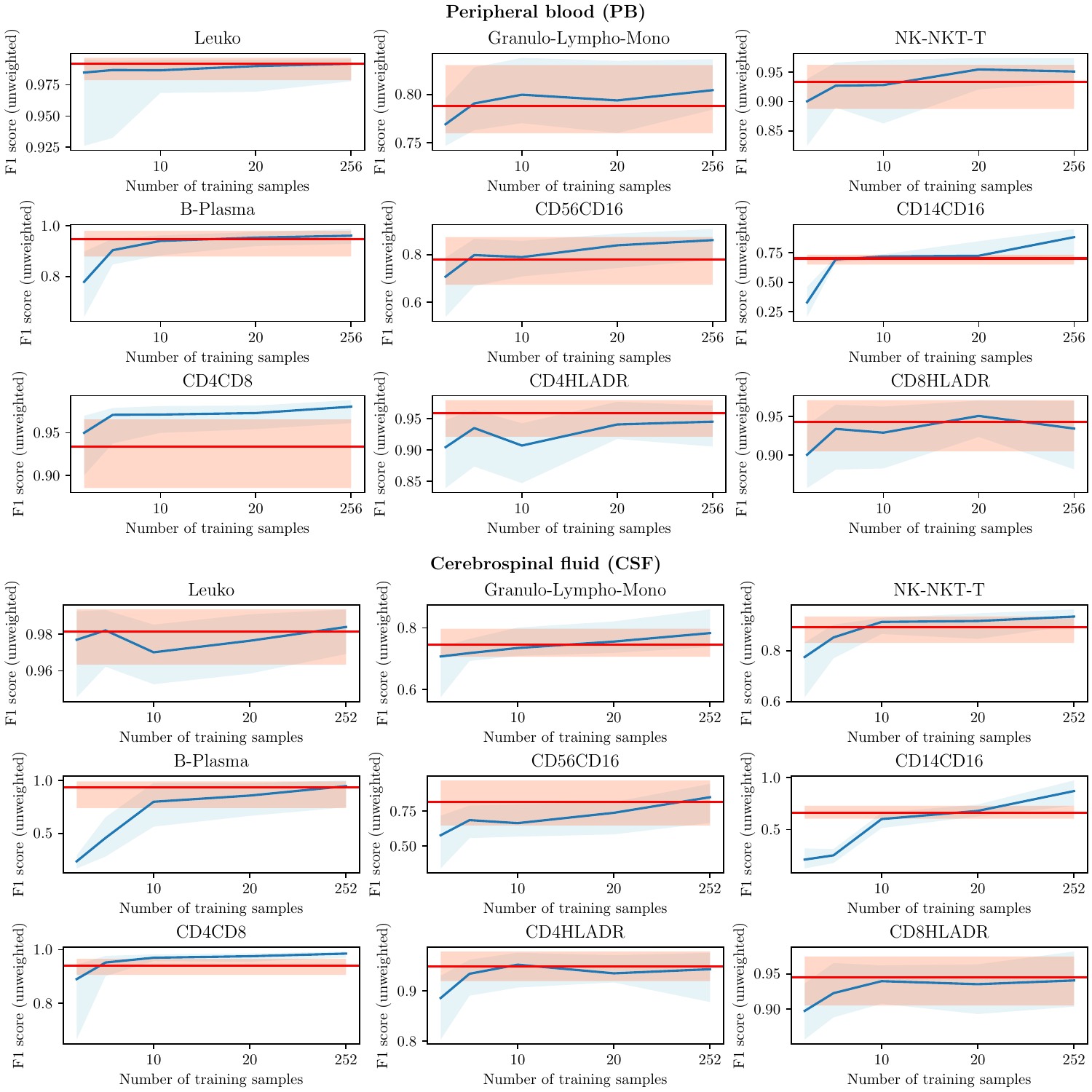}
\caption{Median unweighted F1 score across samples of human experts (red line) and GateNet trained 2, 5, 10, 20 and all training samples (blue line). 25th to 75th percentile depicted as shading in the respective color.} \label{learncurves}
\end{figure}

Community adoption of GateNet will likely depend on its ease of implementation for new cell types. We therefore investigated how many training samples suffice to achieve robust gating. We investigated how the number of training samples impacted performance with a 5-fold cross-validation, which was repeated with 2, 5, 10 and 20 training samples. The training samples were randomly selected from the respective training fold.

In consequence of the previous subsection, we softened the performance threshold in terms of the unweighted F1 score beyond which we define that human expert level gating is achieved to the 25th percentile of human experts.

As shown in Figure \ref{learncurves}, after training with 2 samples the median unweighted F1 score of 7 out of 18 subdatasets lay above the 25th percentile of human experts. With only 10 training samples 17 out of 18 subdatasets met (e.g. B-plasma) this human level performance with the exception of the CD4HLADR subdataset of the PB samples. In many cases, training with 10 or 20 samples yielded similar performance as training with all 200+ samples. This illustrates GateNet’s steep learning curve and provides evidence that it can be used with datasets as small as 10 samples. This holds true for samples with under 1000 events per sample such as the CSF samples of the CD8HLADR subdataset with only 440 events per sample.

\section{Materials and Methods}

\subsection{Datasets}

This study uses the RheumaFlow dataset from the Department of Neurology of the University Hospital Münster and two dataset published during the FlowCAP I (Critical Assessment of Population Identification Methods) challenge\cite{flowcap}. All datasets contain flow cytometry data as well as the expert gating i.e. cell labels for each event.

The RheumaFlow datasets comprises measurements from 55 patients with rheumatic diseases consisting of 64 peripheral blood (PB) and 63 cerebrospinal fluid (CSF) samples. Each sample was gated by four different experts, independently from each other. The events of the RheumaFlow dataset are gated hierarchically, first into Leukocytes and non-Leukocytes (Rest) and subsequently into subpopulations of Leukocytes (see Table \ref{hierarchy} and Fig. \ref{gating_strategy}). Further details about the data acquisition can be found in \cite{gro21}.

% Cells from the CSF were analyzed in parallel to EDTA-blood samples. Therefore, CSF was spun down at 290g for 15min. The supernatant was removed and CSF cells were re-suspended in parallel to 100µl EDTA-blood in 1ml VersaLyse (Beckman Coulter) to lyse erythrocytes. Following 10min of incubation, cells were washed twice with flow cytometry buffer (FC buffer: PBS, 2\% heat inactivated FCS, 2mM EDTA; 300g 4min). Cells were stained in 100µl FC buffer supplemented with fluorochrome-conjugated antibodies directed against CD3 (UCHT1), CD4 (13B8.2), CD8 (B9.11), CD14 (RM052), CD16 (3G8), CD19 (J3-119), CD45 (J33), CD56 (N901), CD138 (B-A38), and HLA-DR (Immu-357) (all Beckman Coulter, clones in brackets). After washing, cells were re-suspended in FC buffer, 20µl flow count fluorospheres (Beckman Coulter) were added, and samples were acquired using a Navios flow cytometer (Beckman Coulter).

The Normal Donors (ND) dataset was published within the context of the FlowCAP I challenges (see Section \ref{secA1} for details). The events were gated non-hierarchically (for a detailed description see \cite{flowcap}).

\begin{table}[ht]
\begin{center}
\begin{minipage}{\textwidth}
\caption{Hierarchical gating strategy present in RheumaFlow dataset. To simply referencing each stage has a subdataset name.}\label{hierarchy}
\begin{tabular*}{\textwidth}{@{\extracolsep{\fill}}lcc@{\extracolsep{\fill}}}
\toprule%
Subdataset & Parentpopulation & Subpopulations \\
\midrule
Leuko & - & Leukokytes, Rest \\ 
Granulo-Lympho-Mono & Leukokytes & Granulocytes, Lymphocytes, Monocytes \\ 
NK-NKT-T & Lymphocytes & NKT cells, NK cells, T cells \\ 
B-Plasma & Lymphocytes & B cells, Plasma cells \\ 
CD56CD16 & NK cells & CD5$6^+$, CD5$6^{dim}$CD1$6^+$ Rest \\
CD14CD16 & Monocytes & CD1$4^+$CD1$6^+$, CD1$4^+$CD1$6^-$, CD1$4^-$CD1$6^+$ \\ 
CD4CD8 & T cells & CD$4^+$ T cells, CD$8^+$ T cells \\
CD4HDLAR & CD$4^+$ T cells & CD$4^+$ T cells \\
CD8HDLAR & CD$8^+$ T cells & CD$8^+$ T cells \\
\botrule
\end{tabular*}
\end{minipage}
\end{center}
\end{table}

As the hierarchical gating of the RheumaFlow samples is done in separate gates, each gate can be defined as a separate dataset. Table \ref{datasets} shows the number of samples, events per sample, markers (forward and side scatter included), cell types and the share of minority class events for each dataset.

\begin{table}[ht]
\begin{center}
\begin{minipage}{\textwidth}
\caption{Number of samples $n_s$, number of events per sample $n_e$, number of markers $n_m$, number of cell types $n_c$ and proportion of events per sample which are labeled as the respective minority class $p_{mc}$. The RheumaFlow subdatasets are grouped into peripheral blood (PB) and cerebrospinal fluid (CSF) samples. With respect to the number of events per sample $n_e$ and proportion of minority class $p_{mc}$ the mean (with the standard deviation) of these measures are shown.}\label{datasets}
\begin{tabular*}{\textwidth}{@{\extracolsep{\fill}}lcccccc@{\extracolsep{\fill}}}
\toprule%
Dataset & $n_s$ & $n_e$ & $n_m$ & $n_c$ & $p_{mc}$ \\
\midrule
Leuko (PB) & 64 & $120000\pm110000$ & 12 & 2 & $20.0\pm15.0$ \\ 
Granulo-Lympho-Mono (PB) & 64 & $77000\pm34000$ & 12 & 4 & $4.6\pm2.8$ \\ 
NK-NKT-T (PB) & 64 & $12000\pm11000$ & 12 & 4 & $0.28\pm0.4$ \\ 
B-Plasma (PB) & 64 & $12000\pm11000$ & 12 & 3 & $0.023\pm0.052$ \\ 
CD56CD16 (PB) & 64 & $1600\pm3300$ & 12 & 3 & $0.14\pm0.14$ \\ 
CD14CD16 (PB) & 64 & $5100\pm4500$ & 12 & 4 & $0.32\pm0.33$ \\ 
CD4CD8 (PB) & 64 & $7500\pm6500$ & 12 & 3 & $1.6\pm1.8$ \\ 
CD4HDLAR (PB) & 64 & $5300\pm4600$ & 12 & 2 & $0.54\pm0.41$ \\ 
CD8HDLAR (PB) & 64 & $1800\pm2100$ & 12 & 2 & $0.42\pm0.69$ \\ 
Leuko (CSF) & 63 & $19000\pm22000$ & 12 & 2 & $18.0\pm11.0$ \\ 
Granulo-Lympho-Mono (CSF) & 63 & $5100\pm13000$ & 12 & 4 & $1.5\pm1.5$ \\ 
NK-NKT-T (CSF) & 63 & $2500\pm3900$ & 12 & 4 & $0.18\pm0.2$ \\ 
B-Plasma (CSF) & 63 & $2500\pm3900$ & 12 & 3 & $0.062\pm0.29$ \\ 
CD56CD16 (CSF) & 63 & $100\pm260$ & 12 & 3 & $0.14\pm0.58$ \\ 
CD14CD16 (CSF) & 63 & $600\pm1100$ & 12 & 4 & $0.045\pm0.12$ \\ 
CD4CD8 (CSF) & 63 & $2200\pm3500$ & 12 & 3 & $2.6\pm2.5$ \\ 
CD4HDLAR (CSF) & 63 & $1600\pm3000$ & 12 & 2 & $1.7\pm2.0$ \\ 
CD8HDLAR (CSF) & 63 & $440\pm490$ & 12 & 2 & $1.1\pm1.2$ \\
Normal Donors & 30 & $59000\pm8000$ & 12 & 8 & $0.074\pm0.021$ \\ 
\botrule
\end{tabular*}
\end{minipage}
\end{center}
\end{table}

\subsection{Model architecture}

The central feature of the GateNet architecture is its ability to process multiple events of a sample, in the following referred to as context events, alongside the single event which should be gated. Utilizing these context events, GateNet can form a latent representation of measurement variance caused by the batch effect (Fig. \ref{batch_effect}) across samples. The single event measurement is processed in conjunction with this latent representation, forming a gating prediction which is corrected for the batch effect.

\begin{figure}[ht]
\includegraphics[width=\textwidth]{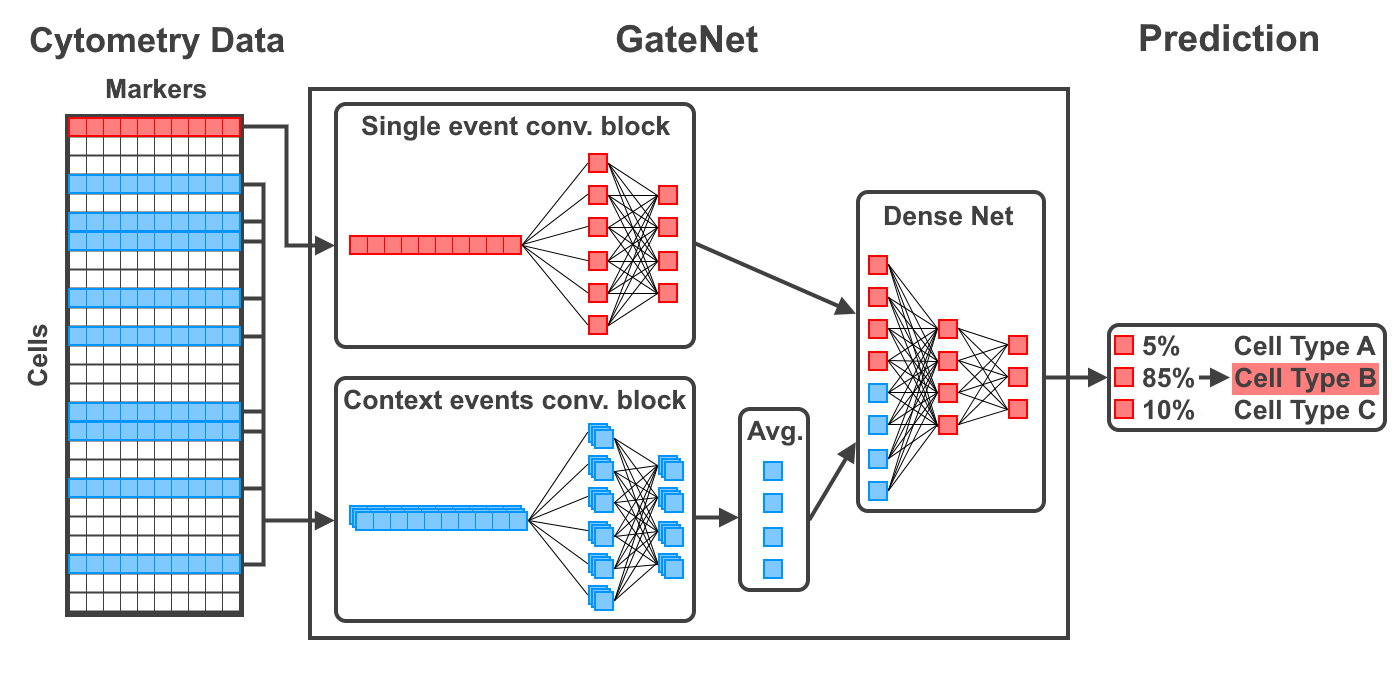}
\caption{Schematic representation of the GateNet architecture.} \label{gatenet}
\end{figure}

GateNet consists of three building blocks: A single event block, a context block and a classification head (Fig. \ref{gatenet}). The architecture of the single event block and the context block is inspired by \cite{hu20} and based on 1D convolutional neural networks (CNN)\cite{cnn}. Both these blocks consist of concatenated modules which comprise one convolutional layer, a batch norm\cite{batchnorm} and a ReLU activation function. The single event block consists of three convolutional modules with 1024, 512 and 256 filters while the context event block contains two modules with 64 and 48 filters. In the context block, these convolutional modules are succeeded by a 1D average pooling layer with a kernel size equal to the number of context events. This effectively implements a simple averaging across the output of all context events. The outputs of the single event block and the context block are then fed into the classification head which is a dense neural network with one hidden layer with 32 nodes. The hidden layer consists of a batch normalization followed by a ReLU activation function. In the final layer a batch normalization is succeeded by a softmax activation function which outputs the predicted probability of each possible cell type for the inputted single event. The model is implemented in PyTorch\cite{pytorch}.\footnote[0]{Implementation at https://github.com/wwu-mmll/gatenet}

\subsection{Training Procedure}
\label{ssec:procedure}

During training, each event is processed accompanied by 1,000 respective events from the same sample. The random sampling of these context events is repeated every time a single event is inputted to the network to avoid overfitting to a certain subset of context events.

Model training is done using the Adam optimizer\cite{adam} and 1cycle learning rate scheduling\cite{onecycle} implemented in the fastai package\cite{fastai}. The 1cycle learning rate policy maximizes convergence by increasing the learning rate to a maximum, here set to 0.002, during the beginning of the training and decreasing it gradually afterwards.

The batch size is set unconventionally high to 1,024 events to reduce computation time since training is done with up to 8,000,000 events. The model is trained until either 5,000 batch iterations, or 10 epochs have been reached. In Section \ref{ssec:small_data}, the batch size is reduced adaptively such that a minimum of 50 batch iterations is reached within 10 epochs. 

\subsection{Imbalanced data methods}
\label{ssec:imbalance}

To counteract poor predictive performance for rare cell classes, the loss function is weighted according to \cite{cui19} with the beta parameter set to 0.99. In addition, the sampling of training events is weighted according to the same logic with a beta of 0.999. Finally, the focal version of the cross-entropy loss is used\cite{lin20}. In focal losses, easy-to-classify samples are down-weights with the gamma parameter which is set to 5 in this work. Those parameter settings were used across all experiments, as the model performance proved to be insensitive to marginal parameter tuning (Supplement Fig. \ref{hparams}).

\subsection{Evaluation}
\label{ssec:eval}

In this work, the model is evaluated in a more challenging, practical fashion compared to the evaluation method used during the FlowCAP I (Critical Assessment of Population Identification Methods)\cite{flowcap}. During the FlowCAP I challenge, 25\% of events of each sample were used for model training and evaluation was done on the remaining 75\% of events of the same sample. Unfortunately, this method of data splitting does not reflect most practical use cases. In practice, gating on new totally unseen samples would be the use case. Furthermore, this data splitting leads to overestimated model performances since the models do not have to correct for measurement variance across samples caused by the batch effect (Supplement Fig. \ref{flowcap}).

Addressing this issue, we use a data splitting scheme which reflects the practical use case with training and validation done with events stemming from different sets of samples (Supplement Fig. \ref{flowcap_split}). In addition, 5-fold cross-validation is used to ensure that the models are not overfitted to a certain subset of validation samples. Finally, performance is evaluated using the F1 score which is the harmonic mean of precision and recall. In the FlowCAP I the total F1 score is a weighted sum of the F1 scores for each cell class. The weights are proportional to the respective population size of the cell class. As pointed out in \cite{web16}, this weighting scheme underrepresents cell classes with small population sizes. Therefore, we report both weighted and unweighted F1 scores.

\section{Discussion}

Here, we developed, implemented and validated a fully automated approach to flow cytometry gating which enables high-accuracy, reliable gating of novel unseen samples. Its performance could be confirmed in an independent, public dataset from the FlowCap I challenge\cite{flowcap}. Importantly, GateNet not only performed comparable to human experts, but matched the consensus of four expert rates more reliably than each human expert. We attribute this to the fact that - unlike human raters who are restricted to sequential 2-dimensional representations - GateNet incorporates the measurements of all markers to achieve an optimal separation of populations across all samples in the training data. Therefore, it can capture the underlying populations with high precision and improve on the gating presented during training (Fig. \ref{3d_gate}). In addition, GateNet training is very data efficient requiring few training samples (Fig. \ref{teaser} and \ref{learncurves}).

In contrast to current automation approaches using clustering, which is prone to instability\cite{mel17}, GateNet enabled stable gating across samples. Furthermore, manual adjustment of algorithm parameters, which is needed for clustering, is rendered obsolete by GateNet as it produces stable gating with a single set of hyperparameters. In addition, the adjustment of GateNet via training samples is much less ambiguous and requires less knowledge about the algorithm compared to adjustment via hyperparameter settings.

In summary, these results imply that GateNet enables the standardized application of gating strategies in challenging settings such as cross-center studies with thousands of samples measured with different cytometers. Therefore, sharing pretrained GateNet models will aid in standardizing immunophenotyping as required by immunology research consortia\cite{cos17, cos19}. GateNet can be trained with very few samples (Fig. \ref{teaser} and \ref{3d_gate}). This capability enhances its utility for ensuring dependable gating in both high-throughput laboratories and smaller studies, such as those in clinical settings, where sample sizes may be $\approx$10.

Generally, GateNet will foster flow cytometry methods in general since it is not only more reliable but also more cost efficient and faster than manual human gating. The implementation utilizes the compute capability of graphics processing units (GPU) allowing to gate a typical sample containing 10.000 to 100.000 events in 0.15 to 1.5 seconds. Besides this speedup, publicly shared GateNet models could lower the entry barrier to flow cytometry analysis for researchers and clinicians with less expertise in gating and advance the use of flow cytometry methods.

Finally, training GateNet could be used in conjunction with existing dimensionality reduction and clustering methods to advance the precision of gating. As shown in Figure \ref{teaser}, reliable gating even with rare cell populations, is possible with only four training samples if those training samples are precisely gated. This meticulous gating of training samples could be facilitated with existing dimensionality reduction and clustering methods. Vice versa, the identification of novel cell populations via existing clustering methods could be supported by GateNet which could help to distinguish clusters caused by random fluctuations from clusters which occur reliably across samples.

% \bmhead{Acknowledgments}

% Acknowledgments are not compulsory. Where included they should be brief. Grant or contribution numbers may be acknowledged.

\section*{Declarations}
\subsection*{Funding}

This work was funded by the German Research Foundation (DFG grants HA7070/2-2, HA7070/3, HA7070/4 to TH) and the Interdisciplinary Center for Clinical Research (IZKF) of the medical faculty of Münster (grants Dan3/012/17 to UD and MzH 3/020/20 to TH and GMzH). G.M.z.H. was supported by grants from the Deutsche Forschungsgemeinschaft (DFG): grant number ME4050/12-1, the Heisenberg program of the DFG (grant number ME4050/13-1). M.Heming was supported by the Interdisciplinary Center for Clinical Research (IZKF) of the medical faculty of Münster (grant SEED/016/21 to M.Heming).

\begin{appendices}

\section{FlowCAP I}\label{secA1}

In Supplementary Figure \ref{flowcap} forward scatter (FSC) versus side scatter (SSC) scatter plots are shown for three example samples for each of the five datasets published during the FlowCAP I. The distribution of gating labels have to be consistent across samples in the respective datasets to enable training and validation on separate sets of samples. 

\begin{figure}[ht]
\includegraphics[width=\textwidth]{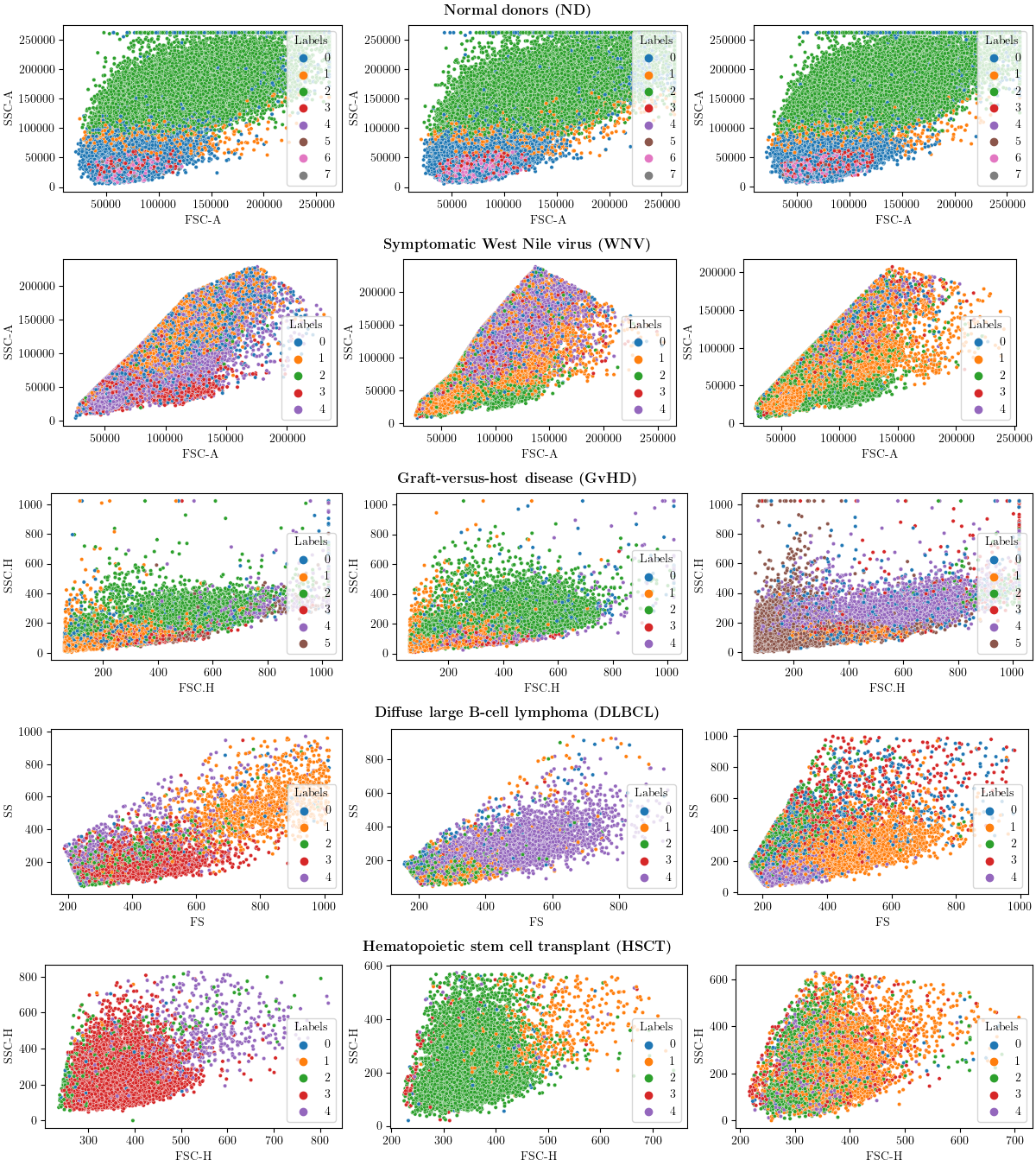}
\caption{Ground truth labels of three example samples from each of the five datasets.} \label{flowcap}
\end{figure}

Unfortunately, only the Normal donors (ND) dataset events are labeled consistent across their respective samples. In the other datasets, the labels of populations are set differently in each sample. In the Hematopoietic stem cell transplant (HSCT) dataset for example, the population showing the lowest FSC intensity is labeled “2” in the left sample but labeled “3” in the middle and right sample.

This inconsistency across samples does not harm the performance if training and validation is done with events of the same sample as it was done during the FlowCAP I challenges. However, when models are validated on samples they were not trained on, labels must be consistent across samples.

% \begin{figure}[H]
% \includegraphics[width=\textwidth]{figures/wnv.png}
% \caption{Labels of three example samples of the West Nile Virus (WNV) dataset before (top) and after label correction (bottom).} \label{wnv}
% \end{figure}

% As shown in Figure \ref{wnv}, we corrected the WNV dataset manually to restore consistent labeling across samples.

\begin{figure}[ht]
\includegraphics[width=\textwidth]{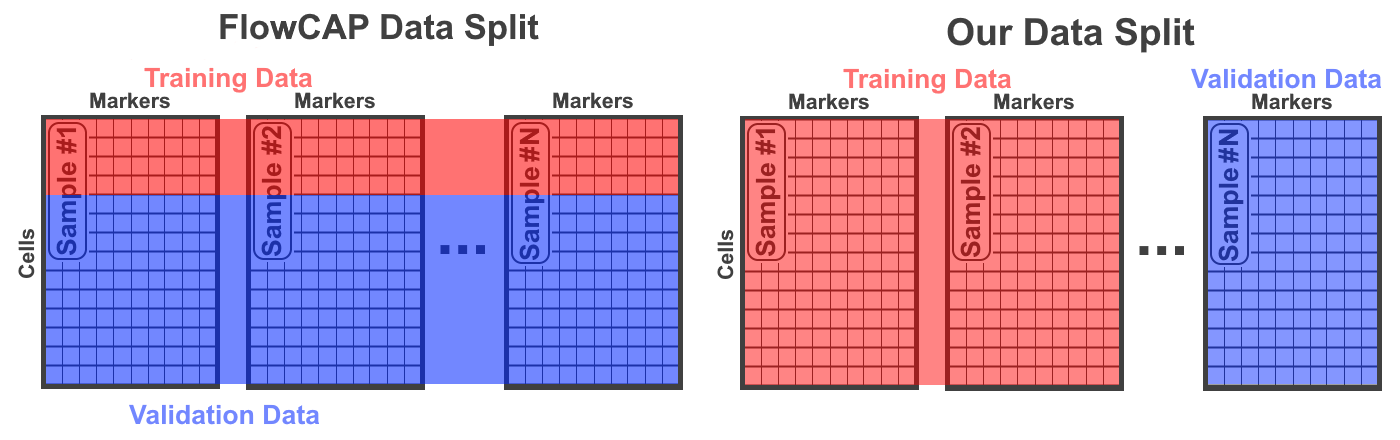}
\caption{Data split used during the FlowCAP 1 (left), and data split used during this analysis (right).} \label{flowcap_split}
\end{figure}

\section{Comparison to DGCyTOF and DeepCyTOF}\label{secA2}

The unrealistic data split used during FlowCAP 1 (Supplement Fig. \ref{flowcap_split} left) results in overestimated performance metrics. As shown in Supplementary Figure \ref{other_nets}, existing models DGCyTOF and DeepCyTOF which proclaim to perform with a weighted F1 score of $>$0.98 do not manage to gate reliably with our more realistic data split (Supplement Fig. \ref{flowcap_split} right).

Both DGCyTOF and DeepCyTOF were applied using the optimizer, learning rate schedule and batch size described in Section \ref{ssec:procedure} with otherwise original hyperparameters.

\begin{figure}[ht]
\includegraphics[width=\textwidth]{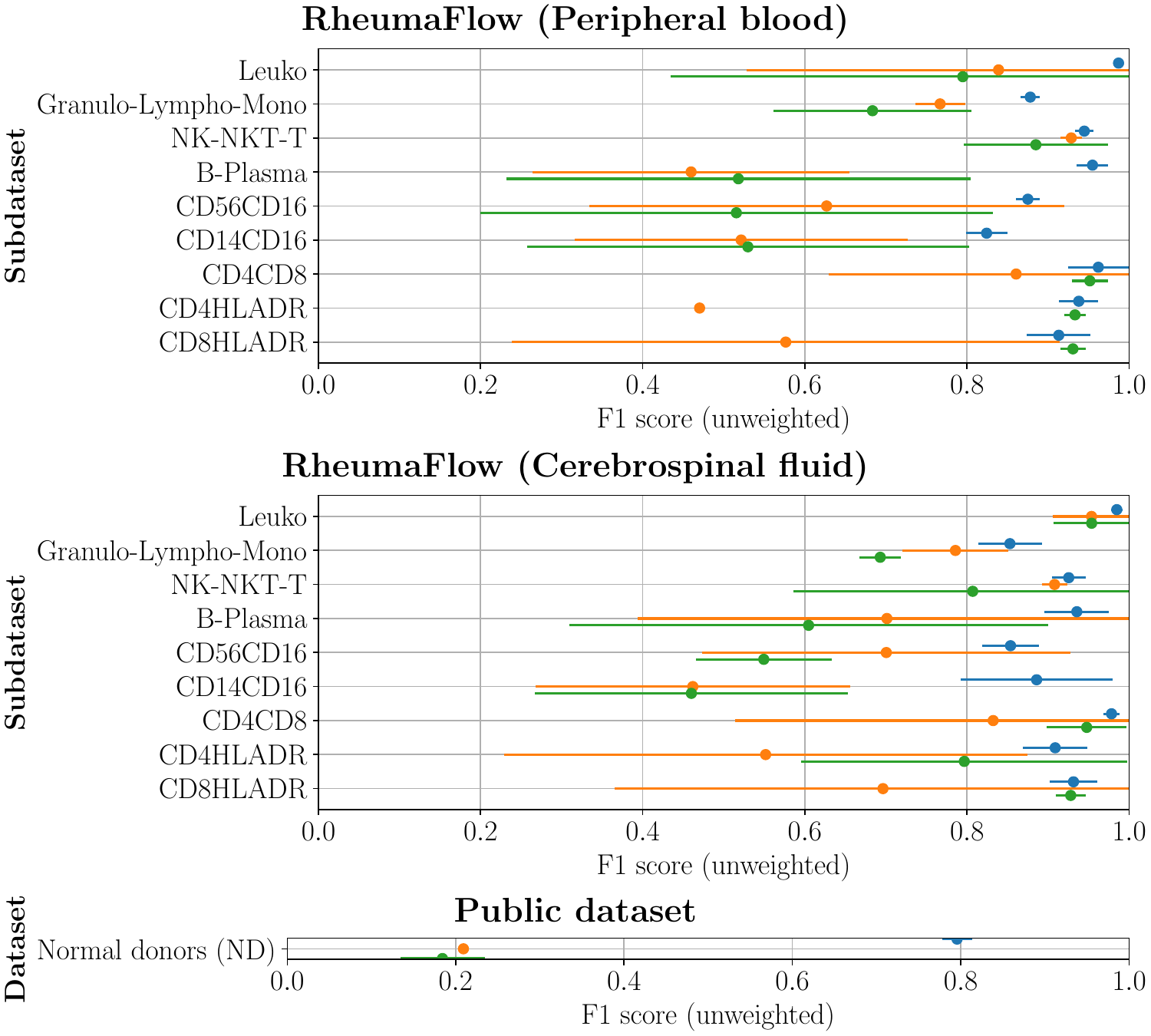}
\caption{Unweighted F1 score of GateNet, DGCyTOF and DeepCyTOF obtained during 5-fold cross-validation in the validation sets. Standard deviations across folds are shown as error bars.} \label{other_nets}
\end{figure}

\section{RheumaFlow gating strategy}\label{secA3}

\begin{figure}[H]
\includegraphics[width=\textwidth]{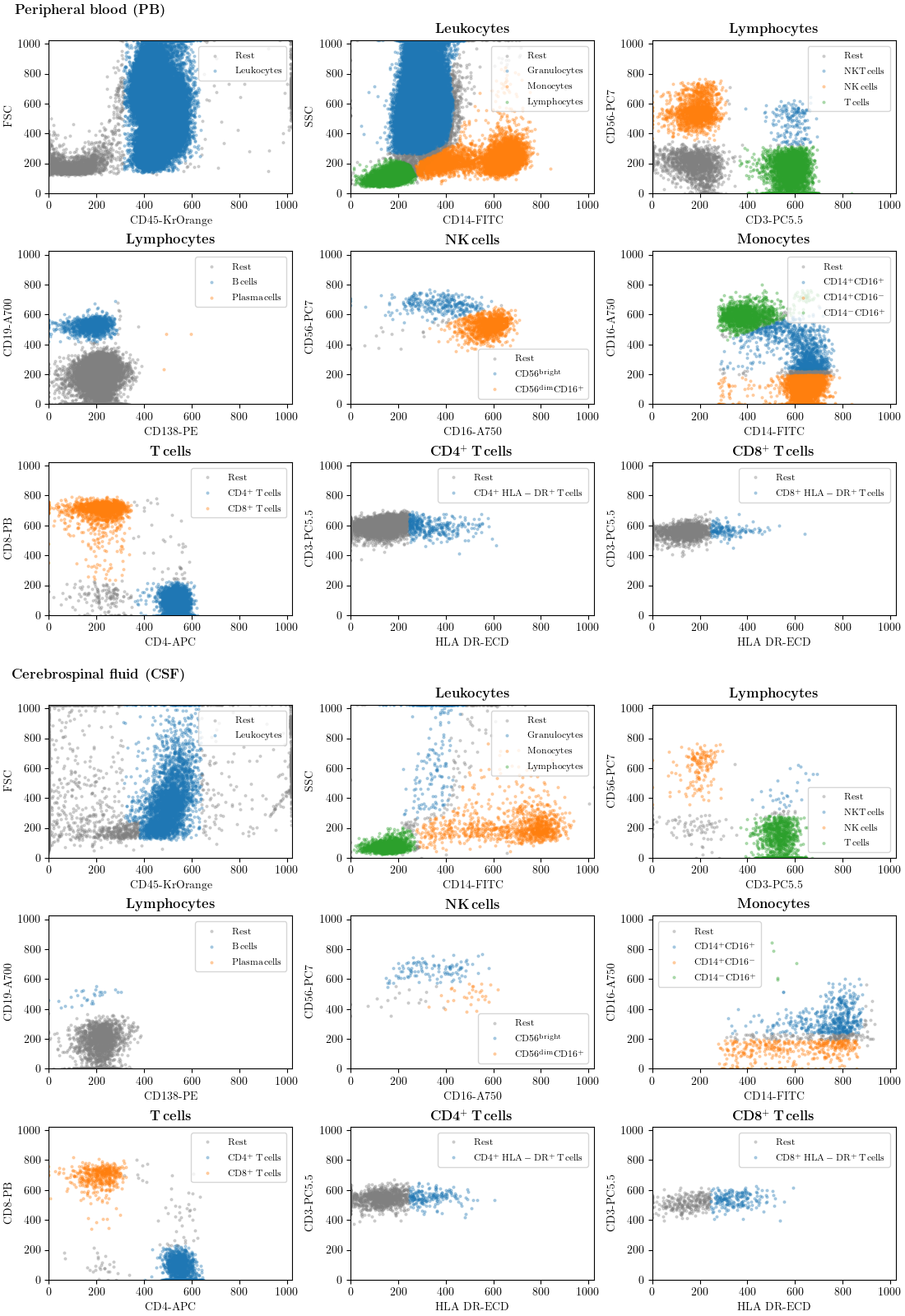}
\caption{Manual gating strategy in the RheumaFlow dataset.} \label{gating_strategy}
\end{figure}

% \begin{figure}[H]
% \includegraphics[width=0.99\textwidth]{figures/worst_samples.png}
% \caption{Worst samples.} \label{worst}
% \end{figure}

\section{Additional Results}\label{secA4}

\begin{table}[h]
\begin{center}
\begin{minipage}{274pt}
\caption{Mean and standard deviation of performance metrics obtained during 5-fold cross-validation in the validation sets. The standard deviation (Std.) across folds is given in parentheses.}\label{results}
\begin{tabular}{@{}lll@{}}
\toprule
Dataset & F1 score (weighted) & F1 score (unweighted) \\
\midrule
Leuko (PB)              &      $0.994\pm0.004$ &        $0.987\pm0.005$ \\
Granulo-Lympho-Mono (PB) &      $0.994\pm0.001$ &        $0.878\pm0.012$ \\
NK-NKT-T (PB)            &      $0.986\pm0.007$ &        $0.945\pm0.011$ \\
B-Plasma (PB)           &      $0.997\pm0.002$ &        $0.955\pm0.019$ \\
CD56CD16 (PB)           &      $0.969\pm0.003$ &        $0.875\pm0.014$ \\
CD14CD16 (PB)           &      $0.962\pm0.021$ &        $0.824\pm0.026$ \\
CD4CD8 (PB)             &      $0.991\pm0.007$ &        $0.962\pm0.038$ \\
CD4HLADR (PB)         &      $0.979\pm0.007$ &        $0.938\pm0.024$ \\
CD8HLADR (PB)         &      $0.950\pm0.019$ &        $0.913\pm0.039$ \\
Leuko (CSF)              &      $0.994\pm0.002$ &        $0.985\pm0.007$ \\
Granulo-Lympho-Mono (CSF) &      $0.965\pm0.019$ &        $0.853\pm0.039$ \\
NK-NKT-T (CSF)            &      $0.991\pm0.003$ &        $0.925\pm0.021$ \\
B-Plasma (CSF)            &      $0.998\pm0.001$ &        $0.935\pm0.040$ \\
CD56CD16 (CSF)           &      $0.910\pm0.034$ &        $0.854\pm0.035$ \\
CD14CD16 (CSF)           &      $0.955\pm0.031$ &        $0.886\pm0.094$ \\
CD4CD8 (CSF)             &      $0.993\pm0.004$ &        $0.978\pm0.010$ \\
CD4HLADR (CSF)          &      $0.950\pm0.021$ &        $0.909\pm0.039$ \\
CD8HLADR (CSF)         &      $0.945\pm0.023$ &        $0.931\pm0.029$ \\
Normal donors (ND)    &      $0.936\pm0.010$ &        $0.796\pm0.016$ \\
% West Nile Virus (WNV) &      $0.886\pm0.034$ &        $0.726\pm0.053$ \\
\botrule
\end{tabular}
\end{minipage}
\end{center}
\end{table}

\section{Parameter Settings}\label{secA4}

\begin{figure}[H]
\includegraphics[width=\textwidth]{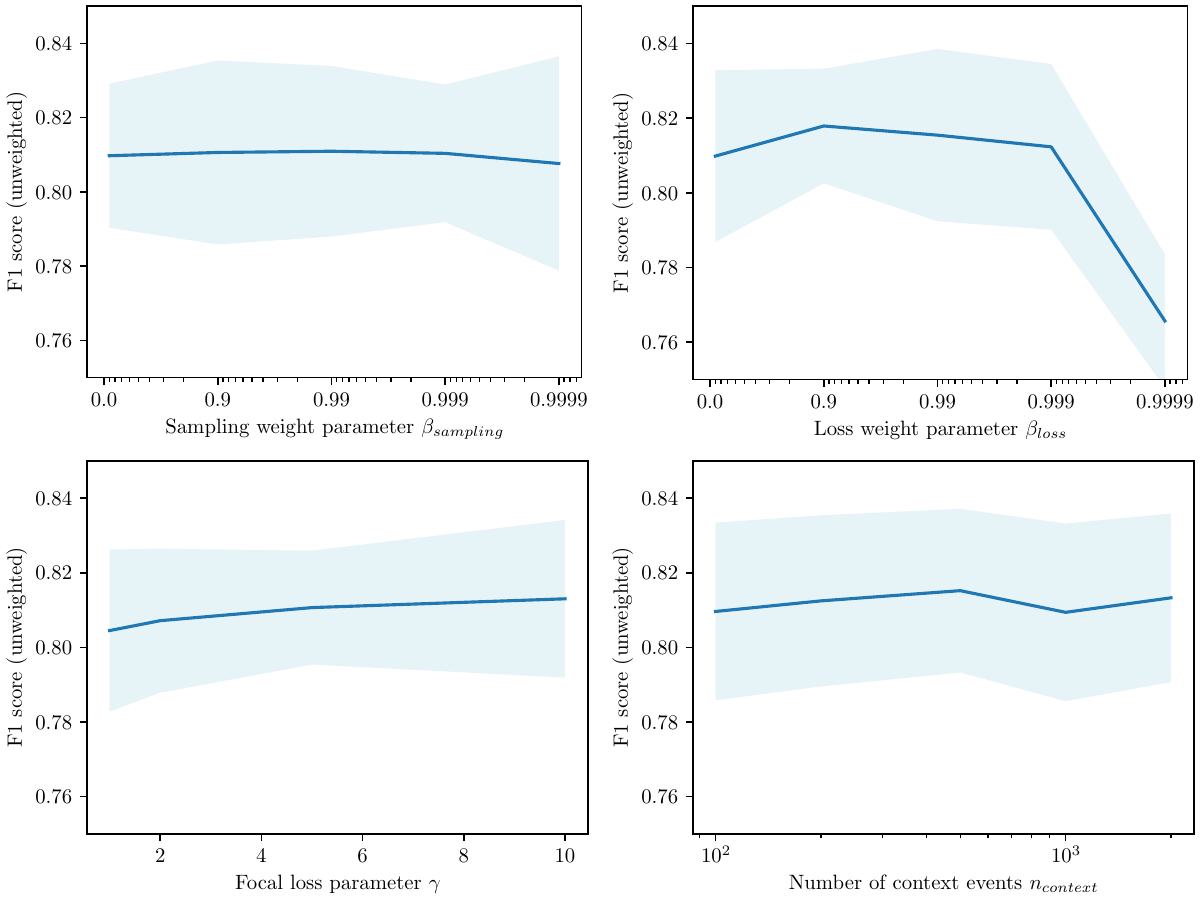}
\caption{Unweighted F1 score obtained during 5-fold cross-validation in the Normal Donors dataset in dependence of model parameter settings. The standard deviation of F1 scores across validation folds is displayed as shading.} \label{hparams}
\end{figure}

\end{appendices}


%\bibliography{bibliography.bib}% common bib file

\begin{thebibliography}{26}
\ifx \bisbn   \undefined \def \bisbn  #1{ISBN #1}\fi
\ifx \binits  \undefined \def \binits#1{#1}\fi
\ifx \bauthor  \undefined \def \bauthor#1{#1}\fi
\ifx \batitle  \undefined \def \batitle#1{#1}\fi
\ifx \bjtitle  \undefined \def \bjtitle#1{#1}\fi
\ifx \bvolume  \undefined \def \bvolume#1{\textbf{#1}}\fi
\ifx \byear  \undefined \def \byear#1{#1}\fi
\ifx \bissue  \undefined \def \bissue#1{#1}\fi
\ifx \bfpage  \undefined \def \bfpage#1{#1}\fi
\ifx \blpage  \undefined \def \blpage #1{#1}\fi
\ifx \burl  \undefined \def \burl#1{\textsf{#1}}\fi
\ifx \doiurl  \undefined \def \doiurl#1{\url{https://doi.org/#1}}\fi
\ifx \betal  \undefined \def \betal{\textit{et al.}}\fi
\ifx \binstitute  \undefined \def \binstitute#1{#1}\fi
\ifx \binstitutionaled  \undefined \def \binstitutionaled#1{#1}\fi
\ifx \bctitle  \undefined \def \bctitle#1{#1}\fi
\ifx \beditor  \undefined \def \beditor#1{#1}\fi
\ifx \bpublisher  \undefined \def \bpublisher#1{#1}\fi
\ifx \bbtitle  \undefined \def \bbtitle#1{#1}\fi
\ifx \bedition  \undefined \def \bedition#1{#1}\fi
\ifx \bseriesno  \undefined \def \bseriesno#1{#1}\fi
\ifx \blocation  \undefined \def \blocation#1{#1}\fi
\ifx \bsertitle  \undefined \def \bsertitle#1{#1}\fi
\ifx \bsnm \undefined \def \bsnm#1{#1}\fi
\ifx \bsuffix \undefined \def \bsuffix#1{#1}\fi
\ifx \bparticle \undefined \def \bparticle#1{#1}\fi
\ifx \barticle \undefined \def \barticle#1{#1}\fi
\bibcommenthead
\ifx \bconfdate \undefined \def \bconfdate #1{#1}\fi
\ifx \botherref \undefined \def \botherref #1{#1}\fi
\ifx \url \undefined \def \url#1{\textsf{#1}}\fi
\ifx \bchapter \undefined \def \bchapter#1{#1}\fi
\ifx \bbook \undefined \def \bbook#1{#1}\fi
\ifx \bcomment \undefined \def \bcomment#1{#1}\fi
\ifx \oauthor \undefined \def \oauthor#1{#1}\fi
\ifx \citeauthoryear \undefined \def \citeauthoryear#1{#1}\fi
\ifx \endbibitem  \undefined \def \endbibitem {}\fi
\ifx \bconflocation  \undefined \def \bconflocation#1{#1}\fi
\ifx \arxivurl  \undefined \def \arxivurl#1{\textsf{#1}}\fi
\csname PreBibitemsHook\endcsname

%%% 1
\bibitem{cra08}
\begin{barticle}
\bauthor{\bsnm{Craig}, \binits{F.E.}},
\bauthor{\bsnm{Foon}, \binits{K.A.}}:
\batitle{{Flow cytometric immunophenotyping for hematologic neoplasms}}.
\bjtitle{Blood}
\bvolume{111}(\bissue{8}),
\bfpage{3941}--\blpage{3967}
(\byear{2008})
{\href{https://arxiv.org/abs/https://ashpublications.org/blood/article-pdf/111/8/3941/1299467/zh800808003941.pdf}{{https://ashpublications.org/blood/article-pdf/111/8/3941/1299467/zh800808003941.pdf}}}.
\doiurl{10.1182/blood-2007-11-120535}
\end{barticle}
\endbibitem

%%% 2
\bibitem{bau00}
\begin{barticle}
\bauthor{\bsnm{Baumgarth}, \binits{N.}},
\bauthor{\bsnm{Roederer}, \binits{M.}}:
\batitle{A practical approach to multicolor flow cytometry for
  immunophenotyping}.
\bjtitle{Journal of Immunological Methods}
\bvolume{243}(\bissue{1}),
\bfpage{77}--\blpage{97}
(\byear{2000}).
\doiurl{10.1016/S0022-1759(00)00229-5}.
\bcomment{Flow Cytometry}
\end{barticle}
\endbibitem

%%% 3
\bibitem{gra20}
\begin{botherref}
\oauthor{\bsnm{Grant}, \binits{R.}},
\oauthor{\bsnm{Coopman}, \binits{K.}},
\oauthor{\bsnm{Medcalf}, \binits{N.}},
\oauthor{\bsnm{Silva-Gomes}, \binits{S.}},
\oauthor{\bsnm{Campbell}, \binits{J.J.}},
\oauthor{\bsnm{Kara}, \binits{B.}},
\oauthor{\bsnm{Braybrook}, \binits{J.}},
\oauthor{\bsnm{Petzing}, \binits{J.}}:
Understanding the contribution of operator measurement variability within flow
  cytometry data analysis for quality control of cell and gene therapy
  manufacturing.
Measurement: Journal of the International Measurement Confederation
\textbf{150}
(2020).
\doiurl{10.1016/j.measurement.2019.106998}
\end{botherref}
\endbibitem

%%% 4
\bibitem{gra21}
\begin{botherref}
\oauthor{\bsnm{Grant}, \binits{R.}},
\oauthor{\bsnm{Coopman}, \binits{K.}},
\oauthor{\bsnm{Medcalf}, \binits{N.}},
\oauthor{\bsnm{Silva-Gomes}, \binits{S.}},
\oauthor{\bsnm{Campbell}, \binits{J.J.}},
\oauthor{\bsnm{Kara}, \binits{B.}},
\oauthor{\bsnm{Braybrook}, \binits{J.}},
\oauthor{\bsnm{Petzing}, \binits{J.}}:
Quantifying operator subjectivity within flow cytometry data analysis as a
  source of measurement uncertainty and the impact of experience on results.
PDA journal of pharmaceutical science and technology
\textbf{75}
(2021).
\doiurl{10.5731/pdajpst.2019.011213}
\end{botherref}
\endbibitem

%%% 5
\bibitem{fin16}
\begin{botherref}
\oauthor{\bsnm{Finak}, \binits{G.}},
\oauthor{\bsnm{Langweiler}, \binits{M.}},
\oauthor{\bsnm{Jaimes}, \binits{M.}},
\oauthor{\bsnm{Malek}, \binits{M.}},
\oauthor{\bsnm{Taghiyar}, \binits{J.}},
\oauthor{\bsnm{Korin}, \binits{Y.}},
\oauthor{\bsnm{Raddassi}, \binits{K.}},
\oauthor{\bsnm{Devine}, \binits{L.}},
\oauthor{\bsnm{Obermoser}, \binits{G.}},
\oauthor{\bsnm{Pekalski}, \binits{M.L.}},
\oauthor{\bsnm{Pontikos}, \binits{N.}},
\oauthor{\bsnm{Diaz}, \binits{A.}},
\oauthor{\bsnm{Heck}, \binits{S.}},
\oauthor{\bsnm{Villanova}, \binits{F.}},
\oauthor{\bsnm{Terrazzini}, \binits{N.}},
\oauthor{\bsnm{Kern}, \binits{F.}},
\oauthor{\bsnm{Qian}, \binits{Y.}},
\oauthor{\bsnm{Stanton}, \binits{R.}},
\oauthor{\bsnm{Wang}, \binits{K.}},
\oauthor{\bsnm{Brandes}, \binits{A.}},
\oauthor{\bsnm{Ramey}, \binits{J.}},
\oauthor{\bsnm{Aghaeepour}, \binits{N.}},
\oauthor{\bsnm{Mosmann}, \binits{T.}},
\oauthor{\bsnm{Scheuermann}, \binits{R.H.}},
\oauthor{\bsnm{Reed}, \binits{E.}},
\oauthor{\bsnm{Palucka}, \binits{K.}},
\oauthor{\bsnm{Pascual}, \binits{V.}},
\oauthor{\bsnm{Blomberg}, \binits{B.B.}},
\oauthor{\bsnm{Nestle}, \binits{F.}},
\oauthor{\bsnm{Nussenblatt}, \binits{R.B.}},
\oauthor{\bsnm{Brinkman}, \binits{R.R.}},
\oauthor{\bsnm{Gottardo}, \binits{R.}},
\oauthor{\bsnm{Maecker}, \binits{H.}},
\oauthor{\bsnm{McCoy}, \binits{J.P.}}:
Standardizing flow cytometry immunophenotyping analysis from the human
  immunophenotyping consortium.
Scientific Reports
\textbf{6}
(2016).
\doiurl{10.1038/srep20686}
\end{botherref}
\endbibitem

%%% 6
\bibitem{qui07}
\begin{barticle}
\bauthor{\bsnm{Quinn}, \binits{J.}},
\bauthor{\bsnm{Fisher}, \binits{P.W.}},
\bauthor{\bsnm{Capocasale}, \binits{R.}},
\bauthor{\bsnm{Achuthanandam}, \binits{R.}},
\bauthor{\bsnm{Kam}, \binits{M.}},
\bauthor{\bsnm{Bugelski}, \binits{P.}},
\bauthor{\bsnm{Hrebien}, \binits{L.}}:
\batitle{A statistical pattern recognition approach for determining cellular
  viability and lineage phenotype in cultured cells and murine bone marrow}.
\bjtitle{Cytometry. Part A : the journal of the International Society for
  Analytical Cytology}
\bvolume{71},
\bfpage{612}--\blpage{24}
(\byear{2007}).
\doiurl{10.1002/cyto.a.20416}
\end{barticle}
\endbibitem

%%% 7
\bibitem{sch19}
\begin{botherref}
\oauthor{\bsnm{Schuyler}, \binits{R.P.}},
\oauthor{\bsnm{Jackson}, \binits{C.}},
\oauthor{\bsnm{Garcia-Perez}, \binits{J.E.}},
\oauthor{\bsnm{Baxter}, \binits{R.M.}},
\oauthor{\bsnm{Ogolla}, \binits{S.}},
\oauthor{\bsnm{Rochford}, \binits{R.}},
\oauthor{\bsnm{Ghosh}, \binits{D.}},
\oauthor{\bsnm{Rudra}, \binits{P.}},
\oauthor{\bsnm{Hsieh}, \binits{E.W.Y.}}:
Minimizing batch effects in mass cytometry data.
Frontiers in Immunology
\textbf{10}
(2019).
\doiurl{10.3389/fimmu.2019.02367}
\end{botherref}
\endbibitem

%%% 8
\bibitem{sha17}
\begin{barticle}
\bauthor{\bsnm{Shaham}, \binits{U.}},
\bauthor{\bsnm{Stanton}, \binits{K.P.}},
\bauthor{\bsnm{Zhao}, \binits{J.}},
\bauthor{\bsnm{Li}, \binits{H.}},
\bauthor{\bsnm{Raddassi}, \binits{K.}},
\bauthor{\bsnm{Montgomery}, \binits{R.}},
\bauthor{\bsnm{Kluger}, \binits{Y.}}:
\batitle{{Removal of batch effects using distribution-matching residual
  networks}}.
\bjtitle{Bioinformatics}
\bvolume{33}(\bissue{16}),
\bfpage{2539}--\blpage{2546}
(\byear{2017})
{\href{https://arxiv.org/abs/https://academic.oup.com/bioinformatics/article-pdf/33/16/2539/25163647/btx196.pdf}{{https://academic.oup.com/bioinformatics/article-pdf/33/16/2539/25163647/btx196.pdf}}}.
\doiurl{10.1093/bioinformatics/btx196}
\end{barticle}
\endbibitem

%%% 9
\bibitem{mel17}
\begin{botherref}
\oauthor{\bsnm{Melchiotti}, \binits{R.}},
\oauthor{\bsnm{Gracio}, \binits{F.}},
\oauthor{\bsnm{Kordasti}, \binits{S.}},
\oauthor{\bsnm{Todd}, \binits{A.K.}},
\oauthor{\bparticle{de} \bsnm{Rinaldis}, \binits{E.}}:
Cluster stability in the analysis of mass cytometry data.
Cytometry Part A
\textbf{91}
(2017).
\doiurl{10.1002/cyto.a.23001}
\end{botherref}
\endbibitem

%%% 10
\bibitem{che21}
\begin{barticle}
\bauthor{\bsnm{Cheung}, \binits{M.}},
\bauthor{\bsnm{Campbell}, \binits{J.J.}},
\bauthor{\bsnm{Whitby}, \binits{L.}},
\bauthor{\bsnm{Thomas}, \binits{R.J.}},
\bauthor{\bsnm{Braybrook}, \binits{J.}},
\bauthor{\bsnm{Petzing}, \binits{J.}}:
\batitle{Current trends in flow cytometry automated data analysis software}.
\bjtitle{Cytometry Part A}
\bvolume{99}(\bissue{10}),
\bfpage{1007}--\blpage{1021}
(\byear{2021})
{\href{https://arxiv.org/abs/https://onlinelibrary.wiley.com/doi/pdf/10.1002/cyto.a.24320}{{https://onlinelibrary.wiley.com/doi/pdf/10.1002/cyto.a.24320}}}.
\doiurl{10.1002/cyto.a.24320}
\end{barticle}
\endbibitem

%%% 11
\bibitem{che22}
\begin{barticle}
\bauthor{\bsnm{Cheng}, \binits{L.}},
\bauthor{\bsnm{Karkhanis}, \binits{P.}},
\bauthor{\bsnm{Gokbag}, \binits{B.}},
\bauthor{\bsnm{Liu}, \binits{Y.}},
\bauthor{\bsnm{Li}, \binits{L.}}:
\batitle{Dgcytof: Deep learning with graphic cluster visualization to predict
  cell types of single cell mass cytometry data}.
\bjtitle{PLOS Computational Biology}
\bvolume{18}(\bissue{4}),
\bfpage{1}--\blpage{22}
(\byear{2022}).
\doiurl{10.1371/journal.pcbi.1008885}
\end{barticle}
\endbibitem

%%% 12
\bibitem{li17}
\begin{barticle}
\bauthor{\bsnm{Li}, \binits{H.}},
\bauthor{\bsnm{Shaham}, \binits{U.}},
\bauthor{\bsnm{Stanton}, \binits{K.P.}},
\bauthor{\bsnm{Yao}, \binits{Y.}},
\bauthor{\bsnm{Montgomery}, \binits{R.R.}},
\bauthor{\bsnm{Kluger}, \binits{Y.}}:
\batitle{{Gating mass cytometry data by deep learning}}.
\bjtitle{Bioinformatics}
\bvolume{33}(\bissue{21}),
\bfpage{3423}--\blpage{3430}
(\byear{2017})
{\href{https://arxiv.org/abs/https://academic.oup.com/bioinformatics/article-pdf/33/21/3423/25166108/btx448.pdf}{{https://academic.oup.com/bioinformatics/article-pdf/33/21/3423/25166108/btx448.pdf}}}.
\doiurl{10.1093/bioinformatics/btx448}
\end{barticle}
\endbibitem

%%% 13
\bibitem{flowcap}
\begin{barticle}
\bauthor{\bsnm{Aghaeepour}, \binits{N.}},
\bauthor{\bsnm{Finak}, \binits{G.}},
\bauthor{\bsnm{Hoos}, \binits{H.}},
\bauthor{\bsnm{Mosmann}, \binits{T.}},
\bauthor{\bsnm{Brinkman}, \binits{R.}},
\bauthor{\bsnm{Gottardo}, \binits{R.}},
\bauthor{\bsnm{Scheuermann}, \binits{R.}}:
\batitle{Critical assessment of automated flow cytometry data analysis
  techniques}.
\bjtitle{Nature Methods}
\bvolume{10},
\bfpage{445}--\blpage{445}
(\byear{2013}).
\doiurl{10.1038/nmeth0513-445c}
\end{barticle}
\endbibitem

%%% 14
\bibitem{web16}
\begin{botherref}
\oauthor{\bsnm{Weber}, \binits{L.M.}},
\oauthor{\bsnm{Robinson}, \binits{M.D.}}:
Comparison of clustering methods for high-dimensional single-cell flow and mass
  cytometry data.
Cytometry Part A
\textbf{89}
(2016).
\doiurl{10.1002/cyto.a.23030}
\end{botherref}
\endbibitem

%%% 15
\bibitem{gro21}
\begin{barticle}
\bauthor{\bsnm{Gross}, \binits{C.C.}},
\bauthor{\bsnm{Schulte-Mecklenbeck}, \binits{A.}},
\bauthor{\bsnm{Madireddy}, \binits{L.}},
\bauthor{\bsnm{Pawlitzki}, \binits{M.}},
\bauthor{\bsnm{Strippel}, \binits{C.}},
\bauthor{\bsnm{Räuber}, \binits{S.}},
\bauthor{\bsnm{Krämer}, \binits{J.}},
\bauthor{\bsnm{Rolfes}, \binits{L.}},
\bauthor{\bsnm{Ruck}, \binits{T.}},
\bauthor{\bsnm{Beuker}, \binits{C.}},
\bauthor{\bsnm{Schmidt-Pogoda}, \binits{A.}},
\bauthor{\bsnm{Lohmann}, \binits{L.}},
\bauthor{\bsnm{Schneider-Hohendorf}, \binits{T.}},
\bauthor{\bsnm{Hahn}, \binits{T.}},
\bauthor{\bsnm{Schwab}, \binits{N.}},
\bauthor{\bsnm{Minnerup}, \binits{J.}},
\bauthor{\bsnm{Melzer}, \binits{N.}},
\bauthor{\bsnm{Klotz}, \binits{L.}},
\bauthor{\bsnm{Meuth}, \binits{S.G.}},
\bauthor{\bparticle{Meyer~zu} \bsnm{Hörste}, \binits{G.}},
\bauthor{\bsnm{Baranzini}, \binits{S.E.}},
\bauthor{\bsnm{Wiendl}, \binits{H.}}:
\batitle{{Classification of neurological diseases using multi-dimensional CSF
  analysis}}.
\bjtitle{Brain}
\bvolume{144}(\bissue{9}),
\bfpage{2625}--\blpage{2634}
(\byear{2021})
{\href{https://arxiv.org/abs/https://academic.oup.com/brain/article-pdf/144/9/2625/40880184/awab147.pdf}{{https://academic.oup.com/brain/article-pdf/144/9/2625/40880184/awab147.pdf}}}.
\doiurl{10.1093/brain/awab147}
\end{barticle}
\endbibitem

%%% 16
\bibitem{hu20}
\begin{barticle}
\bauthor{\bsnm{Hu}, \binits{Z.}},
\bauthor{\bsnm{Tang}, \binits{A.}},
\bauthor{\bsnm{Singh}, \binits{J.}},
\bauthor{\bsnm{Bhattacharya}, \binits{S.}},
\bauthor{\bsnm{Butte}, \binits{A.J.}}:
\batitle{A robust and interpretable end-to-end deep learning model for
  cytometry data}.
\bjtitle{Proceedings of the National Academy of Sciences}
\bvolume{117}(\bissue{35}),
\bfpage{21373}--\blpage{21380}
(\byear{2020})
{\href{https://arxiv.org/abs/https://www.pnas.org/doi/pdf/10.1073/pnas.2003026117}{{https://www.pnas.org/doi/pdf/10.1073/pnas.2003026117}}}.
\doiurl{10.1073/pnas.2003026117}
\end{barticle}
\endbibitem

%%% 17
\bibitem{cnn}
\begin{barticle}
\bauthor{\bsnm{Krizhevsky}, \binits{A.}},
\bauthor{\bsnm{Sutskever}, \binits{I.}},
\bauthor{\bsnm{Hinton}, \binits{G.E.}}:
\batitle{Imagenet classification with deep convolutional neural networks}.
\bjtitle{Commun. ACM}
\bvolume{60}(\bissue{6}),
\bfpage{84}--\blpage{90}
(\byear{2017}).
\doiurl{10.1145/3065386}
\end{barticle}
\endbibitem

%%% 18
\bibitem{batchnorm}
\begin{barticle}
\bauthor{\bsnm{Ioffe}, \binits{S.}},
\bauthor{\bsnm{Szegedy}, \binits{C.}}:
\batitle{Batch normalization: Accelerating deep network training by reducing
  internal covariate shift}.
\bjtitle{32nd International Conference on Machine Learning, ICML 2015}
\bvolume{1},
\bfpage{448}--\blpage{456}
(\byear{2015})
\end{barticle}
\endbibitem

%%% 19
\bibitem{pytorch}
\begin{bchapter}
\bauthor{\bsnm{Paszke}, \binits{A.}},
\bauthor{\bsnm{Gross}, \binits{S.}},
\bauthor{\bsnm{Massa}, \binits{F.}},
\bauthor{\bsnm{Lerer}, \binits{A.}},
\bauthor{\bsnm{Bradbury}, \binits{J.}},
\bauthor{\bsnm{Chanan}, \binits{G.}},
\bauthor{\bsnm{Killeen}, \binits{T.}},
\bauthor{\bsnm{Lin}, \binits{Z.}},
\bauthor{\bsnm{Gimelshein}, \binits{N.}},
\bauthor{\bsnm{Antiga}, \binits{L.}},
\bauthor{\bsnm{Desmaison}, \binits{A.}},
\bauthor{\bsnm{Kopf}, \binits{A.}},
\bauthor{\bsnm{Yang}, \binits{E.}},
\bauthor{\bsnm{DeVito}, \binits{Z.}},
\bauthor{\bsnm{Raison}, \binits{M.}},
\bauthor{\bsnm{Tejani}, \binits{A.}},
\bauthor{\bsnm{Chilamkurthy}, \binits{S.}},
\bauthor{\bsnm{Steiner}, \binits{B.}},
\bauthor{\bsnm{Fang}, \binits{L.}},
\bauthor{\bsnm{Bai}, \binits{J.}},
\bauthor{\bsnm{Chintala}, \binits{S.}}:
\bctitle{Pytorch: An imperative style, high-performance deep learning library}.
In: \beditor{\bsnm{Wallach}, \binits{H.}},
\beditor{\bsnm{Larochelle}, \binits{H.}},
\beditor{\bsnm{Beygelzimer}, \binits{A.}},
\beditor{\bparticle{d\textquotesingle} \bsnm{Alch\'{e}-Buc}, \binits{F.}},
\beditor{\bsnm{Fox}, \binits{E.}},
\beditor{\bsnm{Garnett}, \binits{R.}} (eds.)
\bbtitle{Advances in Neural Information Processing Systems 32},
pp. \bfpage{8024}--\blpage{8035}.
\bpublisher{Curran Associates, Inc.}, \blocation{???}
(\byear{2019}).
\burl{http://papers.neurips.cc/paper/9015-pytorch-an-imperative-style-high-performance-deep-learning-library.pdf}
\end{bchapter}
\endbibitem

%%% 20
\bibitem{adam}
\begin{botherref}
\oauthor{\bsnm{Kingma}, \binits{D.P.}},
\oauthor{\bsnm{Ba}, \binits{J.L.}}:
Adam: A method for stochastic optimization.
3rd International Conference on Learning Representations, ICLR 2015 -
  Conference Track Proceedings
(2015)
\end{botherref}
\endbibitem

%%% 21
\bibitem{onecycle}
\begin{bchapter}
\bauthor{\bsnm{Smith}, \binits{L.N.}}:
\bctitle{Cyclical learning rates for training neural networks},
pp. \bfpage{464}--\blpage{472}.
\bpublisher{Institute of Electrical and Electronics Engineers Inc.},
  \blocation{???}
(\byear{2017}).
\doiurl{10.1109/WACV.2017.58}
\end{bchapter}
\endbibitem

%%% 22
\bibitem{fastai}
\begin{barticle}
\bauthor{\bsnm{Howard}, \binits{J.}},
\bauthor{\bsnm{Gugger}, \binits{S.}}:
\batitle{Fastai: A layered api for deep learning}.
\bjtitle{Information (Switzerland)}
\bvolume{11},
\bfpage{108}
(\byear{2020}).
\doiurl{10.3390/info11020108}
\end{barticle}
\endbibitem

%%% 23
\bibitem{cui19}
\begin{bchapter}
\bauthor{\bsnm{Cui}, \binits{Y.}},
\bauthor{\bsnm{Jia}, \binits{M.}},
\bauthor{\bsnm{Lin}, \binits{T.Y.}},
\bauthor{\bsnm{Song}, \binits{Y.}},
\bauthor{\bsnm{Belongie}, \binits{S.}}:
\bctitle{Class-balanced loss based on effective number of samples},
vol. \bseriesno{2019-June}
(\byear{2019}).
\doiurl{10.1109/CVPR.2019.00949}
\end{bchapter}
\endbibitem

%%% 24
\bibitem{lin20}
\begin{botherref}
\oauthor{\bsnm{Lin}, \binits{T.Y.}},
\oauthor{\bsnm{Goyal}, \binits{P.}},
\oauthor{\bsnm{Girshick}, \binits{R.}},
\oauthor{\bsnm{He}, \binits{K.}},
\oauthor{\bsnm{Dollar}, \binits{P.}}:
Focal loss for dense object detection.
IEEE Transactions on Pattern Analysis and Machine Intelligence
\textbf{42}
(2020).
\doiurl{10.1109/TPAMI.2018.2858826}
\end{botherref}
\endbibitem

%%% 25
\bibitem{cos17}
\begin{barticle}
\bauthor{\bsnm{Cossarizza}, \binits{A.}},
\bauthor{\bsnm{Chang}, \binits{H.-D.}},
\bauthor{\bsnm{Radbruch}, \binits{A.}},
\bauthor{\bsnm{Akdis}, \binits{M.}},
\bauthor{\bsnm{Andrä}, \binits{I.}},
\bauthor{\bsnm{Annunziato}, \binits{F.}},
\bauthor{\bsnm{Bacher}, \binits{P.}},
\bauthor{\bsnm{Barnaba}, \binits{V.}},
\bauthor{\bsnm{Battistini}, \binits{L.}},
\bauthor{\bsnm{Bauer}, \binits{W.M.}},
\bauthor{\bsnm{Baumgart}, \binits{S.}},
\bauthor{\bsnm{Becher}, \binits{B.}},
\bauthor{\bsnm{Beisker}, \binits{W.}},
\bauthor{\bsnm{Berek}, \binits{C.}},
\bauthor{\bsnm{Blanco}, \binits{A.}},
\bauthor{\bsnm{Borsellino}, \binits{G.}},
\bauthor{\bsnm{Boulais}, \binits{P.E.}},
\bauthor{\bsnm{Brinkman}, \binits{R.R.}},
\bauthor{\bsnm{Büscher}, \binits{M.}},
\bauthor{\bsnm{Busch}, \binits{D.H.}},
\bauthor{\bsnm{Bushnell}, \binits{T.P.}},
\bauthor{\bsnm{Cao}, \binits{X.}},
\bauthor{\bsnm{Cavani}, \binits{A.}},
\bauthor{\bsnm{Chattopadhyay}, \binits{P.K.}},
\bauthor{\bsnm{Cheng}, \binits{Q.}},
\bauthor{\bsnm{Chow}, \binits{S.}},
\bauthor{\bsnm{Clerici}, \binits{M.}},
\bauthor{\bsnm{Cooke}, \binits{A.}},
\bauthor{\bsnm{Cosma}, \binits{A.}},
\bauthor{\bsnm{Cosmi}, \binits{L.}},
\bauthor{\bsnm{Cumano}, \binits{A.}},
\bauthor{\bsnm{Dang}, \binits{V.D.}},
\bauthor{\bsnm{Davies}, \binits{D.}},
\bauthor{\bsnm{De~Biasi}, \binits{S.}},
\bauthor{\bsnm{Del~Zotto}, \binits{G.}},
\bauthor{\bsnm{Della~Bella}, \binits{S.}},
\bauthor{\bsnm{Dellabona}, \binits{P.}},
\bauthor{\bsnm{Deniz}, \binits{G.}},
\bauthor{\bsnm{Dessing}, \binits{M.}},
\bauthor{\bsnm{Diefenbach}, \binits{A.}},
\bauthor{\bsnm{Di~Santo}, \binits{J.}},
\bauthor{\bsnm{Dieli}, \binits{F.}},
\bauthor{\bsnm{Dolf}, \binits{A.}},
\bauthor{\bsnm{Donnenberg}, \binits{V.S.}},
\bauthor{\bsnm{Dörner}, \binits{T.}},
\bauthor{\bsnm{Ehrhardt}, \binits{G.R.A.}},
\bauthor{\bsnm{Endl}, \binits{E.}},
\bauthor{\bsnm{Engel}, \binits{P.}},
\bauthor{\bsnm{Engelhardt}, \binits{B.}},
\bauthor{\bsnm{Esser}, \binits{C.}},
\bauthor{\bsnm{Everts}, \binits{B.}},
\bauthor{\bsnm{Dreher}, \binits{A.}},
\bauthor{\bsnm{Falk}, \binits{C.S.}},
\bauthor{\bsnm{Fehniger}, \binits{T.A.}},
\bauthor{\bsnm{Filby}, \binits{A.}},
\bauthor{\bsnm{Fillatreau}, \binits{S.}},
\bauthor{\bsnm{Follo}, \binits{M.}},
\bauthor{\bsnm{Förster}, \binits{I.}},
\bauthor{\bsnm{Foster}, \binits{J.}},
\bauthor{\bsnm{Foulds}, \binits{G.A.}},
\bauthor{\bsnm{Frenette}, \binits{P.S.}},
\bauthor{\bsnm{Galbraith}, \binits{D.}},
\bauthor{\bsnm{Garbi}, \binits{N.}},
\bauthor{\bsnm{García-Godoy}, \binits{M.D.}},
\bauthor{\bsnm{Geginat}, \binits{J.}},
\bauthor{\bsnm{Ghoreschi}, \binits{K.}},
\bauthor{\bsnm{Gibellini}, \binits{L.}},
\bauthor{\bsnm{Goettlinger}, \binits{C.}},
\bauthor{\bsnm{Goodyear}, \binits{C.S.}},
\bauthor{\bsnm{Gori}, \binits{A.}},
\bauthor{\bsnm{Grogan}, \binits{J.}},
\bauthor{\bsnm{Gross}, \binits{M.}},
\bauthor{\bsnm{Grützkau}, \binits{A.}},
\bauthor{\bsnm{Grummitt}, \binits{D.}},
\bauthor{\bsnm{Hahn}, \binits{J.}},
\bauthor{\bsnm{Hammer}, \binits{Q.}},
\bauthor{\bsnm{Hauser}, \binits{A.E.}},
\bauthor{\bsnm{Haviland}, \binits{D.L.}},
\bauthor{\bsnm{Hedley}, \binits{D.}},
\bauthor{\bsnm{Herrera}, \binits{G.}},
\bauthor{\bsnm{Herrmann}, \binits{M.}},
\bauthor{\bsnm{Hiepe}, \binits{F.}},
\bauthor{\bsnm{Holland}, \binits{T.}},
\bauthor{\bsnm{Hombrink}, \binits{P.}},
\bauthor{\bsnm{Houston}, \binits{J.P.}},
\bauthor{\bsnm{Hoyer}, \binits{B.F.}},
\bauthor{\bsnm{Huang}, \binits{B.}},
\bauthor{\bsnm{Hunter}, \binits{C.A.}},
\bauthor{\bsnm{Iannone}, \binits{A.}},
\bauthor{\bsnm{Jäck}, \binits{H.-M.}},
\bauthor{\bsnm{Jávega}, \binits{B.}},
\bauthor{\bsnm{Jonjic}, \binits{S.}},
\bauthor{\bsnm{Juelke}, \binits{K.}},
\bauthor{\bsnm{Jung}, \binits{S.}},
\bauthor{\bsnm{Kaiser}, \binits{T.}},
\bauthor{\bsnm{Kalina}, \binits{T.}},
\bauthor{\bsnm{Keller}, \binits{B.}},
\bauthor{\bsnm{Khan}, \binits{S.}},
\bauthor{\bsnm{Kienhöfer}, \binits{D.}},
\bauthor{\bsnm{Kroneis}, \binits{T.}},
\bauthor{\bsnm{Kunkel}, \binits{D.}},
\bauthor{\bsnm{Kurts}, \binits{C.}},
\bauthor{\bsnm{Kvistborg}, \binits{P.}},
\bauthor{\bsnm{Lannigan}, \binits{J.}},
\bauthor{\bsnm{Lantz}, \binits{O.}},
\bauthor{\bsnm{Larbi}, \binits{A.}},
\bauthor{\bsnm{LeibundGut-Landmann}, \binits{S.}},
\bauthor{\bsnm{Leipold}, \binits{M.D.}},
\bauthor{\bsnm{Levings}, \binits{M.K.}},
\bauthor{\bsnm{Litwin}, \binits{V.}},
\bauthor{\bsnm{Liu}, \binits{Y.}},
\bauthor{\bsnm{Lohoff}, \binits{M.}},
\bauthor{\bsnm{Lombardi}, \binits{G.}},
\bauthor{\bsnm{Lopez}, \binits{L.}},
\bauthor{\bsnm{Lovett-Racke}, \binits{A.}},
\bauthor{\bsnm{Lubberts}, \binits{E.}},
\bauthor{\bsnm{Ludewig}, \binits{B.}},
\bauthor{\bsnm{Lugli}, \binits{E.}},
\bauthor{\bsnm{Maecker}, \binits{H.T.}},
\bauthor{\bsnm{Martrus}, \binits{G.}},
\bauthor{\bsnm{Matarese}, \binits{G.}},
\bauthor{\bsnm{Maueröder}, \binits{C.}},
\bauthor{\bsnm{McGrath}, \binits{M.}},
\bauthor{\bsnm{McInnes}, \binits{I.}},
\bauthor{\bsnm{Mei}, \binits{H.E.}},
\bauthor{\bsnm{Melchers}, \binits{F.}},
\bauthor{\bsnm{Melzer}, \binits{S.}},
\bauthor{\bsnm{Mielenz}, \binits{D.}},
\bauthor{\bsnm{Mills}, \binits{K.}},
\bauthor{\bsnm{Mirrer}, \binits{D.}},
\bauthor{\bsnm{Mjösberg}, \binits{J.}},
\bauthor{\bsnm{Moore}, \binits{J.}},
\bauthor{\bsnm{Moran}, \binits{B.}},
\bauthor{\bsnm{Moretta}, \binits{A.}},
\bauthor{\bsnm{Moretta}, \binits{L.}},
\bauthor{\bsnm{Mosmann}, \binits{T.R.}},
\bauthor{\bsnm{Müller}, \binits{S.}},
\bauthor{\bsnm{Müller}, \binits{W.}},
\bauthor{\bsnm{Münz}, \binits{C.}},
\bauthor{\bsnm{Multhoff}, \binits{G.}},
\bauthor{\bsnm{Munoz}, \binits{L.E.}},
\bauthor{\bsnm{Murphy}, \binits{K.M.}},
\bauthor{\bsnm{Nakayama}, \binits{T.}},
\bauthor{\bsnm{Nasi}, \binits{M.}},
\bauthor{\bsnm{Neudörfl}, \binits{C.}},
\bauthor{\bsnm{Nolan}, \binits{J.}},
\bauthor{\bsnm{Nourshargh}, \binits{S.}},
\bauthor{\bsnm{O'Connor}, \binits{J.-E.}},
\bauthor{\bsnm{Ouyang}, \binits{W.}},
\bauthor{\bsnm{Oxenius}, \binits{A.}},
\bauthor{\bsnm{Palankar}, \binits{R.}},
\bauthor{\bsnm{Panse}, \binits{I.}},
\bauthor{\bsnm{Peterson}, \binits{P.}},
\bauthor{\bsnm{Peth}, \binits{C.}},
\bauthor{\bsnm{Petriz}, \binits{J.}},
\bauthor{\bsnm{Philips}, \binits{D.}},
\bauthor{\bsnm{Pickl}, \binits{W.}},
\bauthor{\bsnm{Piconese}, \binits{S.}},
\bauthor{\bsnm{Pinti}, \binits{M.}},
\bauthor{\bsnm{Pockley}, \binits{A.G.}},
\bauthor{\bsnm{Podolska}, \binits{M.J.}},
\bauthor{\bsnm{Pucillo}, \binits{C.}},
\bauthor{\bsnm{Quataert}, \binits{S.A.}},
\bauthor{\bsnm{Radstake}, \binits{T.R.D.J.}},
\bauthor{\bsnm{Rajwa}, \binits{B.}},
\bauthor{\bsnm{Rebhahn}, \binits{J.A.}},
\bauthor{\bsnm{Recktenwald}, \binits{D.}},
\bauthor{\bsnm{Remmerswaal}, \binits{E.B.M.}},
\bauthor{\bsnm{Rezvani}, \binits{K.}},
\bauthor{\bsnm{Rico}, \binits{L.G.}},
\bauthor{\bsnm{Robinson}, \binits{J.P.}},
\bauthor{\bsnm{Romagnani}, \binits{C.}},
\bauthor{\bsnm{Rubartelli}, \binits{A.}},
\bauthor{\bsnm{Ruckert}, \binits{B.}},
\bauthor{\bsnm{Ruland}, \binits{J.}},
\bauthor{\bsnm{Sakaguchi}, \binits{S.}},
\bauthor{\bsnm{Sala-de-Oyanguren}, \binits{F.}},
\bauthor{\bsnm{Samstag}, \binits{Y.}},
\bauthor{\bsnm{Sanderson}, \binits{S.}},
\bauthor{\bsnm{Sawitzki}, \binits{B.}},
\bauthor{\bsnm{Scheffold}, \binits{A.}},
\bauthor{\bsnm{Schiemann}, \binits{M.}},
\bauthor{\bsnm{Schildberg}, \binits{F.}},
\bauthor{\bsnm{Schimisky}, \binits{E.}},
\bauthor{\bsnm{Schmid}, \binits{S.A.}},
\bauthor{\bsnm{Schmitt}, \binits{S.}},
\bauthor{\bsnm{Schober}, \binits{K.}},
\bauthor{\bsnm{Schüler}, \binits{T.}},
\bauthor{\bsnm{Schulz}, \binits{A.R.}},
\bauthor{\bsnm{Schumacher}, \binits{T.}},
\bauthor{\bsnm{Scotta}, \binits{C.}},
\bauthor{\bsnm{Shankey}, \binits{T.V.}},
\bauthor{\bsnm{Shemer}, \binits{A.}},
\bauthor{\bsnm{Simon}, \binits{A.-K.}},
\bauthor{\bsnm{Spidlen}, \binits{J.}},
\bauthor{\bsnm{Stall}, \binits{A.M.}},
\bauthor{\bsnm{Stark}, \binits{R.}},
\bauthor{\bsnm{Stehle}, \binits{C.}},
\bauthor{\bsnm{Stein}, \binits{M.}},
\bauthor{\bsnm{Steinmetz}, \binits{T.}},
\bauthor{\bsnm{Stockinger}, \binits{H.}},
\bauthor{\bsnm{Takahama}, \binits{Y.}},
\bauthor{\bsnm{Tarnok}, \binits{A.}},
\bauthor{\bsnm{Tian}, \binits{Z.}},
\bauthor{\bsnm{Toldi}, \binits{G.}},
\bauthor{\bsnm{Tornack}, \binits{J.}},
\bauthor{\bsnm{Traggiai}, \binits{E.}},
\bauthor{\bsnm{Trotter}, \binits{J.}},
\bauthor{\bsnm{Ulrich}, \binits{H.}},
\bauthor{\bparticle{van~der} \bsnm{Braber}, \binits{M.}},
\bauthor{\bparticle{van} \bsnm{Lier}, \binits{R.A.W.}},
\bauthor{\bsnm{Veldhoen}, \binits{M.}},
\bauthor{\bsnm{Vento-Asturias}, \binits{S.}},
\bauthor{\bsnm{Vieira}, \binits{P.}},
\bauthor{\bsnm{Voehringer}, \binits{D.}},
\bauthor{\bsnm{Volk}, \binits{H.-D.}},
\bauthor{\bparticle{von} \bsnm{Volkmann}, \binits{K.}},
\bauthor{\bsnm{Waisman}, \binits{A.}},
\bauthor{\bsnm{Walker}, \binits{R.}},
\bauthor{\bsnm{Ward}, \binits{M.D.}},
\bauthor{\bsnm{Warnatz}, \binits{K.}},
\bauthor{\bsnm{Warth}, \binits{S.}},
\bauthor{\bsnm{Watson}, \binits{J.V.}},
\bauthor{\bsnm{Watzl}, \binits{C.}},
\bauthor{\bsnm{Wegener}, \binits{L.}},
\bauthor{\bsnm{Wiedemann}, \binits{A.}},
\bauthor{\bsnm{Wienands}, \binits{J.}},
\bauthor{\bsnm{Willimsky}, \binits{G.}},
\bauthor{\bsnm{Wing}, \binits{J.}},
\bauthor{\bsnm{Wurst}, \binits{P.}},
\bauthor{\bsnm{Yu}, \binits{L.}},
\bauthor{\bsnm{Yue}, \binits{A.}},
\bauthor{\bsnm{Zhang}, \binits{Q.}},
\bauthor{\bsnm{Zhao}, \binits{Y.}},
\bauthor{\bsnm{Ziegler}, \binits{S.}},
\bauthor{\bsnm{Zimmermann}, \binits{J.}}:
\batitle{Guidelines for the use of flow cytometry and cell sorting in
  immunological studies*}.
\bjtitle{European Journal of Immunology}
\bvolume{47}(\bissue{10}),
\bfpage{1584}--\blpage{1797}
(\byear{2017})
{\href{https://arxiv.org/abs/https://onlinelibrary.wiley.com/doi/pdf/10.1002/eji.201646632}{{https://onlinelibrary.wiley.com/doi/pdf/10.1002/eji.201646632}}}.
\doiurl{10.1002/eji.201646632}
\end{barticle}
\endbibitem

%%% 26
\bibitem{cos19}
\begin{barticle}
\bauthor{\bsnm{Cossarizza}, \binits{A.}},
\bauthor{\bsnm{Chang}, \binits{H.-D.}},
\bauthor{\bsnm{Radbruch}, \binits{A.}},
\bauthor{\bsnm{Acs}, \binits{A.}},
\bauthor{\bsnm{Adam}, \binits{D.}},
\bauthor{\bsnm{Adam-Klages}, \binits{S.}},
\bauthor{\bsnm{Agace}, \binits{W.W.}},
\bauthor{\bsnm{Aghaeepour}, \binits{N.}},
\bauthor{\bsnm{Akdis}, \binits{M.}},
\bauthor{\bsnm{Allez}, \binits{M.}},
\bauthor{\bsnm{Almeida}, \binits{L.N.}},
\bauthor{\bsnm{Alvisi}, \binits{G.}},
\bauthor{\bsnm{Anderson}, \binits{G.}},
\bauthor{\bsnm{Andrä}, \binits{I.}},
\bauthor{\bsnm{Annunziato}, \binits{F.}},
\bauthor{\bsnm{Anselmo}, \binits{A.}},
\bauthor{\bsnm{Bacher}, \binits{P.}},
\bauthor{\bsnm{Baldari}, \binits{C.T.}},
\bauthor{\bsnm{Bari}, \binits{S.}},
\bauthor{\bsnm{Barnaba}, \binits{V.}},
\bauthor{\bsnm{Barros-Martins}, \binits{J.}},
\bauthor{\bsnm{Battistini}, \binits{L.}},
\bauthor{\bsnm{Bauer}, \binits{W.}},
\bauthor{\bsnm{Baumgart}, \binits{S.}},
\bauthor{\bsnm{Baumgarth}, \binits{N.}},
\bauthor{\bsnm{Baumjohann}, \binits{D.}},
\bauthor{\bsnm{Baying}, \binits{B.}},
\bauthor{\bsnm{Bebawy}, \binits{M.}},
\bauthor{\bsnm{Becher}, \binits{B.}},
\bauthor{\bsnm{Beisker}, \binits{W.}},
\bauthor{\bsnm{Benes}, \binits{V.}},
\bauthor{\bsnm{Beyaert}, \binits{R.}},
\bauthor{\bsnm{Blanco}, \binits{A.}},
\bauthor{\bsnm{Boardman}, \binits{D.A.}},
\bauthor{\bsnm{Bogdan}, \binits{C.}},
\bauthor{\bsnm{Borger}, \binits{J.G.}},
\bauthor{\bsnm{Borsellino}, \binits{G.}},
\bauthor{\bsnm{Boulais}, \binits{P.E.}},
\bauthor{\bsnm{Bradford}, \binits{J.A.}},
\bauthor{\bsnm{Brenner}, \binits{D.}},
\bauthor{\bsnm{Brinkman}, \binits{R.R.}},
\bauthor{\bsnm{Brooks}, \binits{A.E.S.}},
\bauthor{\bsnm{Busch}, \binits{D.H.}},
\bauthor{\bsnm{Büscher}, \binits{M.}},
\bauthor{\bsnm{Bushnell}, \binits{T.P.}},
\bauthor{\bsnm{Calzetti}, \binits{F.}},
\bauthor{\bsnm{Cameron}, \binits{G.}},
\bauthor{\bsnm{Cammarata}, \binits{I.}},
\bauthor{\bsnm{Cao}, \binits{X.}},
\bauthor{\bsnm{Cardell}, \binits{S.L.}},
\bauthor{\bsnm{Casola}, \binits{S.}},
\bauthor{\bsnm{Cassatella}, \binits{M.A.}},
\bauthor{\bsnm{Cavani}, \binits{A.}},
\bauthor{\bsnm{Celada}, \binits{A.}},
\bauthor{\bsnm{Chatenoud}, \binits{L.}},
\bauthor{\bsnm{Chattopadhyay}, \binits{P.K.}},
\bauthor{\bsnm{Chow}, \binits{S.}},
\bauthor{\bsnm{Christakou}, \binits{E.}},
\bauthor{\bsnm{Čičin-Šain}, \binits{L.}},
\bauthor{\bsnm{Clerici}, \binits{M.}},
\bauthor{\bsnm{Colombo}, \binits{F.S.}},
\bauthor{\bsnm{Cook}, \binits{L.}},
\bauthor{\bsnm{Cooke}, \binits{A.}},
\bauthor{\bsnm{Cooper}, \binits{A.M.}},
\bauthor{\bsnm{Corbett}, \binits{A.J.}},
\bauthor{\bsnm{Cosma}, \binits{A.}},
\bauthor{\bsnm{Cosmi}, \binits{L.}},
\bauthor{\bsnm{Coulie}, \binits{P.G.}},
\bauthor{\bsnm{Cumano}, \binits{A.}},
\bauthor{\bsnm{Cvetkovic}, \binits{L.}},
\bauthor{\bsnm{Dang}, \binits{V.D.}},
\bauthor{\bsnm{Dang-Heine}, \binits{C.}},
\bauthor{\bsnm{Davey}, \binits{M.S.}},
\bauthor{\bsnm{Davies}, \binits{D.}},
\bauthor{\bsnm{De~Biasi}, \binits{S.}},
\bauthor{\bsnm{Del~Zotto}, \binits{G.}},
\bauthor{\bsnm{Dela~Cruz}, \binits{G.V.}},
\bauthor{\bsnm{Delacher}, \binits{M.}},
\bauthor{\bsnm{Della~Bella}, \binits{S.}},
\bauthor{\bsnm{Dellabona}, \binits{P.}},
\bauthor{\bsnm{Deniz}, \binits{G.}},
\bauthor{\bsnm{Dessing}, \binits{M.}},
\bauthor{\bsnm{Di~Santo}, \binits{J.P.}},
\bauthor{\bsnm{Diefenbach}, \binits{A.}},
\bauthor{\bsnm{Dieli}, \binits{F.}},
\bauthor{\bsnm{Dolf}, \binits{A.}},
\bauthor{\bsnm{Dörner}, \binits{T.}},
\bauthor{\bsnm{Dress}, \binits{R.J.}},
\bauthor{\bsnm{Dudziak}, \binits{D.}},
\bauthor{\bsnm{Dustin}, \binits{M.}},
\bauthor{\bsnm{Dutertre}, \binits{C.-A.}},
\bauthor{\bsnm{Ebner}, \binits{F.}},
\bauthor{\bsnm{Eckle}, \binits{S.B.G.}},
\bauthor{\bsnm{Edinger}, \binits{M.}},
\bauthor{\bsnm{Eede}, \binits{P.}},
\bauthor{\bsnm{Ehrhardt}, \binits{G.R.A.}},
\bauthor{\bsnm{Eich}, \binits{M.}},
\bauthor{\bsnm{Engel}, \binits{P.}},
\bauthor{\bsnm{Engelhardt}, \binits{B.}},
\bauthor{\bsnm{Erdei}, \binits{A.}},
\bauthor{\bsnm{Esser}, \binits{C.}},
\bauthor{\bsnm{Everts}, \binits{B.}},
\bauthor{\bsnm{Evrard}, \binits{M.}},
\bauthor{\bsnm{Falk}, \binits{C.S.}},
\bauthor{\bsnm{Fehniger}, \binits{T.A.}},
\bauthor{\bsnm{Felipo-Benavent}, \binits{M.}},
\bauthor{\bsnm{Ferry}, \binits{H.}},
\bauthor{\bsnm{Feuerer}, \binits{M.}},
\bauthor{\bsnm{Filby}, \binits{A.}},
\bauthor{\bsnm{Filkor}, \binits{K.}},
\bauthor{\bsnm{Fillatreau}, \binits{S.}},
\bauthor{\bsnm{Follo}, \binits{M.}},
\bauthor{\bsnm{Förster}, \binits{I.}},
\bauthor{\bsnm{Foster}, \binits{J.}},
\bauthor{\bsnm{Foulds}, \binits{G.A.}},
\bauthor{\bsnm{Frehse}, \binits{B.}},
\bauthor{\bsnm{Frenette}, \binits{P.S.}},
\bauthor{\bsnm{Frischbutter}, \binits{S.}},
\bauthor{\bsnm{Fritzsche}, \binits{W.}},
\bauthor{\bsnm{Galbraith}, \binits{D.W.}},
\bauthor{\bsnm{Gangaev}, \binits{A.}},
\bauthor{\bsnm{Garbi}, \binits{N.}},
\bauthor{\bsnm{Gaudilliere}, \binits{B.}},
\bauthor{\bsnm{Gazzinelli}, \binits{R.T.}},
\bauthor{\bsnm{Geginat}, \binits{J.}},
\bauthor{\bsnm{Gerner}, \binits{W.}},
\bauthor{\bsnm{Gherardin}, \binits{N.A.}},
\bauthor{\bsnm{Ghoreschi}, \binits{K.}},
\bauthor{\bsnm{Gibellini}, \binits{L.}},
\bauthor{\bsnm{Ginhoux}, \binits{F.}},
\bauthor{\bsnm{Goda}, \binits{K.}},
\bauthor{\bsnm{Godfrey}, \binits{D.I.}},
\bauthor{\bsnm{Goettlinger}, \binits{C.}},
\bauthor{\bsnm{González-Navajas}, \binits{J.M.}},
\bauthor{\bsnm{Goodyear}, \binits{C.S.}},
\bauthor{\bsnm{Gori}, \binits{A.}},
\bauthor{\bsnm{Grogan}, \binits{J.L.}},
\bauthor{\bsnm{Grummitt}, \binits{D.}},
\bauthor{\bsnm{Grützkau}, \binits{A.}},
\bauthor{\bsnm{Haftmann}, \binits{C.}},
\bauthor{\bsnm{Hahn}, \binits{J.}},
\bauthor{\bsnm{Hammad}, \binits{H.}},
\bauthor{\bsnm{Hämmerling}, \binits{G.}},
\bauthor{\bsnm{Hansmann}, \binits{L.}},
\bauthor{\bsnm{Hansson}, \binits{G.}},
\bauthor{\bsnm{Harpur}, \binits{C.M.}},
\bauthor{\bsnm{Hartmann}, \binits{S.}},
\bauthor{\bsnm{Hauser}, \binits{A.}},
\bauthor{\bsnm{Hauser}, \binits{A.E.}},
\bauthor{\bsnm{Haviland}, \binits{D.L.}},
\bauthor{\bsnm{Hedley}, \binits{D.}},
\bauthor{\bsnm{Hernández}, \binits{D.C.}},
\bauthor{\bsnm{Herrera}, \binits{G.}},
\bauthor{\bsnm{Herrmann}, \binits{M.}},
\bauthor{\bsnm{Hess}, \binits{C.}},
\bauthor{\bsnm{Höfer}, \binits{T.}},
\bauthor{\bsnm{Hoffmann}, \binits{P.}},
\bauthor{\bsnm{Hogquist}, \binits{K.}},
\bauthor{\bsnm{Holland}, \binits{T.}},
\bauthor{\bsnm{Höllt}, \binits{T.}},
\bauthor{\bsnm{Holmdahl}, \binits{R.}},
\bauthor{\bsnm{Hombrink}, \binits{P.}},
\bauthor{\bsnm{Houston}, \binits{J.P.}},
\bauthor{\bsnm{Hoyer}, \binits{B.F.}},
\bauthor{\bsnm{Huang}, \binits{B.}},
\bauthor{\bsnm{Huang}, \binits{F.-P.}},
\bauthor{\bsnm{Huber}, \binits{J.E.}},
\bauthor{\bsnm{Huehn}, \binits{J.}},
\bauthor{\bsnm{Hundemer}, \binits{M.}},
\bauthor{\bsnm{Hunter}, \binits{C.A.}},
\bauthor{\bsnm{Hwang}, \binits{W.Y.K.}},
\bauthor{\bsnm{Iannone}, \binits{A.}},
\bauthor{\bsnm{Ingelfinger}, \binits{F.}},
\bauthor{\bsnm{Ivison}, \binits{S.M.}},
\bauthor{\bsnm{Jäck}, \binits{H.-M.}},
\bauthor{\bsnm{Jani}, \binits{P.K.}},
\bauthor{\bsnm{Jávega}, \binits{B.}},
\bauthor{\bsnm{Jonjic}, \binits{S.}},
\bauthor{\bsnm{Kaiser}, \binits{T.}},
\bauthor{\bsnm{Kalina}, \binits{T.}},
\bauthor{\bsnm{Kamradt}, \binits{T.}},
\bauthor{\bsnm{Kaufmann}, \binits{S.H.E.}},
\bauthor{\bsnm{Keller}, \binits{B.}},
\bauthor{\bsnm{Ketelaars}, \binits{S.L.C.}},
\bauthor{\bsnm{Khalilnezhad}, \binits{A.}},
\bauthor{\bsnm{Khan}, \binits{S.}},
\bauthor{\bsnm{Kisielow}, \binits{J.}},
\bauthor{\bsnm{Klenerman}, \binits{P.}},
\bauthor{\bsnm{Knopf}, \binits{J.}},
\bauthor{\bsnm{Koay}, \binits{H.-F.}},
\bauthor{\bsnm{Kobow}, \binits{K.}},
\bauthor{\bsnm{Kolls}, \binits{J.K.}},
\bauthor{\bsnm{Kong}, \binits{W.T.}},
\bauthor{\bsnm{Kopf}, \binits{M.}},
\bauthor{\bsnm{Korn}, \binits{T.}},
\bauthor{\bsnm{Kriegsmann}, \binits{K.}},
\bauthor{\bsnm{Kristyanto}, \binits{H.}},
\bauthor{\bsnm{Kroneis}, \binits{T.}},
\bauthor{\bsnm{Krueger}, \binits{A.}},
\bauthor{\bsnm{Kühne}, \binits{J.}},
\bauthor{\bsnm{Kukat}, \binits{C.}},
\bauthor{\bsnm{Kunkel}, \binits{D.}},
\bauthor{\bsnm{Kunze-Schumacher}, \binits{H.}},
\bauthor{\bsnm{Kurosaki}, \binits{T.}},
\bauthor{\bsnm{Kurts}, \binits{C.}},
\bauthor{\bsnm{Kvistborg}, \binits{P.}},
\bauthor{\bsnm{Kwok}, \binits{I.}},
\bauthor{\bsnm{Landry}, \binits{J.}},
\bauthor{\bsnm{Lantz}, \binits{O.}},
\bauthor{\bsnm{Lanuti}, \binits{P.}},
\bauthor{\bsnm{LaRosa}, \binits{F.}},
\bauthor{\bsnm{Lehuen}, \binits{A.}},
\bauthor{\bsnm{LeibundGut-Landmann}, \binits{S.}},
\bauthor{\bsnm{Leipold}, \binits{M.D.}},
\bauthor{\bsnm{Leung}, \binits{L.Y.T.}},
\bauthor{\bsnm{Levings}, \binits{M.K.}},
\bauthor{\bsnm{Lino}, \binits{A.C.}},
\bauthor{\bsnm{Liotta}, \binits{F.}},
\bauthor{\bsnm{Litwin}, \binits{V.}},
\bauthor{\bsnm{Liu}, \binits{Y.}},
\bauthor{\bsnm{Ljunggren}, \binits{H.-G.}},
\bauthor{\bsnm{Lohoff}, \binits{M.}},
\bauthor{\bsnm{Lombardi}, \binits{G.}},
\bauthor{\bsnm{Lopez}, \binits{L.}},
\bauthor{\bsnm{López-Botet}, \binits{M.}},
\bauthor{\bsnm{Lovett-Racke}, \binits{A.E.}},
\bauthor{\bsnm{Lubberts}, \binits{E.}},
\bauthor{\bsnm{Luche}, \binits{H.}},
\bauthor{\bsnm{Ludewig}, \binits{B.}},
\bauthor{\bsnm{Lugli}, \binits{E.}},
\bauthor{\bsnm{Lunemann}, \binits{S.}},
\bauthor{\bsnm{Maecker}, \binits{H.T.}},
\bauthor{\bsnm{Maggi}, \binits{L.}},
\bauthor{\bsnm{Maguire}, \binits{O.}},
\bauthor{\bsnm{Mair}, \binits{F.}},
\bauthor{\bsnm{Mair}, \binits{K.H.}},
\bauthor{\bsnm{Mantovani}, \binits{A.}},
\bauthor{\bsnm{Manz}, \binits{R.A.}},
\bauthor{\bsnm{Marshall}, \binits{A.J.}},
\bauthor{\bsnm{Martínez-Romero}, \binits{A.}},
\bauthor{\bsnm{Martrus}, \binits{G.}},
\bauthor{\bsnm{Marventano}, \binits{I.}},
\bauthor{\bsnm{Maslinski}, \binits{W.}},
\bauthor{\bsnm{Matarese}, \binits{G.}},
\bauthor{\bsnm{Mattioli}, \binits{A.V.}},
\bauthor{\bsnm{Maueröder}, \binits{C.}},
\bauthor{\bsnm{Mazzoni}, \binits{A.}},
\bauthor{\bsnm{McCluskey}, \binits{J.}},
\bauthor{\bsnm{McGrath}, \binits{M.}},
\bauthor{\bsnm{McGuire}, \binits{H.M.}},
\bauthor{\bsnm{McInnes}, \binits{I.B.}},
\bauthor{\bsnm{Mei}, \binits{H.E.}},
\bauthor{\bsnm{Melchers}, \binits{F.}},
\bauthor{\bsnm{Melzer}, \binits{S.}},
\bauthor{\bsnm{Mielenz}, \binits{D.}},
\bauthor{\bsnm{Miller}, \binits{S.D.}},
\bauthor{\bsnm{Mills}, \binits{K.H.G.}},
\bauthor{\bsnm{Minderman}, \binits{H.}},
\bauthor{\bsnm{Mjösberg}, \binits{J.}},
\bauthor{\bsnm{Moore}, \binits{J.}},
\bauthor{\bsnm{Moran}, \binits{B.}},
\bauthor{\bsnm{Moretta}, \binits{L.}},
\bauthor{\bsnm{Mosmann}, \binits{T.R.}},
\bauthor{\bsnm{Müller}, \binits{S.}},
\bauthor{\bsnm{Multhoff}, \binits{G.}},
\bauthor{\bsnm{Muñoz}, \binits{L.E.}},
\bauthor{\bsnm{Münz}, \binits{C.}},
\bauthor{\bsnm{Nakayama}, \binits{T.}},
\bauthor{\bsnm{Nasi}, \binits{M.}},
\bauthor{\bsnm{Neumann}, \binits{K.}},
\bauthor{\bsnm{Ng}, \binits{L.G.}},
\bauthor{\bsnm{Niedobitek}, \binits{A.}},
\bauthor{\bsnm{Nourshargh}, \binits{S.}},
\bauthor{\bsnm{Núñez}, \binits{G.}},
\bauthor{\bsnm{O'Connor}, \binits{J.-E.}},
\bauthor{\bsnm{Ochel}, \binits{A.}},
\bauthor{\bsnm{Oja}, \binits{A.}},
\bauthor{\bsnm{Ordonez}, \binits{D.}},
\bauthor{\bsnm{Orfao}, \binits{A.}},
\bauthor{\bsnm{Orlowski-Oliver}, \binits{E.}},
\bauthor{\bsnm{Ouyang}, \binits{W.}},
\bauthor{\bsnm{Oxenius}, \binits{A.}},
\bauthor{\bsnm{Palankar}, \binits{R.}},
\bauthor{\bsnm{Panse}, \binits{I.}},
\bauthor{\bsnm{Pattanapanyasat}, \binits{K.}},
\bauthor{\bsnm{Paulsen}, \binits{M.}},
\bauthor{\bsnm{Pavlinic}, \binits{D.}},
\bauthor{\bsnm{Penter}, \binits{L.}},
\bauthor{\bsnm{Peterson}, \binits{P.}},
\bauthor{\bsnm{Peth}, \binits{C.}},
\bauthor{\bsnm{Petriz}, \binits{J.}},
\bauthor{\bsnm{Piancone}, \binits{F.}},
\bauthor{\bsnm{Pickl}, \binits{W.F.}},
\bauthor{\bsnm{Piconese}, \binits{S.}},
\bauthor{\bsnm{Pinti}, \binits{M.}},
\bauthor{\bsnm{Pockley}, \binits{A.G.}},
\bauthor{\bsnm{Podolska}, \binits{M.J.}},
\bauthor{\bsnm{Poon}, \binits{Z.}},
\bauthor{\bsnm{Pracht}, \binits{K.}},
\bauthor{\bsnm{Prinz}, \binits{I.}},
\bauthor{\bsnm{Pucillo}, \binits{C.E.M.}},
\bauthor{\bsnm{Quataert}, \binits{S.A.}},
\bauthor{\bsnm{Quatrini}, \binits{L.}},
\bauthor{\bsnm{Quinn}, \binits{K.M.}},
\bauthor{\bsnm{Radbruch}, \binits{H.}},
\bauthor{\bsnm{Radstake}, \binits{T.R.D.J.}},
\bauthor{\bsnm{Rahmig}, \binits{S.}},
\bauthor{\bsnm{Rahn}, \binits{H.-P.}},
\bauthor{\bsnm{Rajwa}, \binits{B.}},
\bauthor{\bsnm{Ravichandran}, \binits{G.}},
\bauthor{\bsnm{Raz}, \binits{Y.}},
\bauthor{\bsnm{Rebhahn}, \binits{J.A.}},
\bauthor{\bsnm{Recktenwald}, \binits{D.}},
\bauthor{\bsnm{Reimer}, \binits{D.}},
\bauthor{\bparticle{Reis~e} \bsnm{Sousa}, \binits{C.}},
\bauthor{\bsnm{Remmerswaal}, \binits{E.B.M.}},
\bauthor{\bsnm{Richter}, \binits{L.}},
\bauthor{\bsnm{Rico}, \binits{L.G.}},
\bauthor{\bsnm{Riddell}, \binits{A.}},
\bauthor{\bsnm{Rieger}, \binits{A.M.}},
\bauthor{\bsnm{Robinson}, \binits{J.P.}},
\bauthor{\bsnm{Romagnani}, \binits{C.}},
\bauthor{\bsnm{Rubartelli}, \binits{A.}},
\bauthor{\bsnm{Ruland}, \binits{J.}},
\bauthor{\bsnm{Saalmüller}, \binits{A.}},
\bauthor{\bsnm{Saeys}, \binits{Y.}},
\bauthor{\bsnm{Saito}, \binits{T.}},
\bauthor{\bsnm{Sakaguchi}, \binits{S.}},
\bauthor{\bsnm{Sala-de-Oyanguren}, \binits{F.}},
\bauthor{\bsnm{Samstag}, \binits{Y.}},
\bauthor{\bsnm{Sanderson}, \binits{S.}},
\bauthor{\bsnm{Sandrock}, \binits{I.}},
\bauthor{\bsnm{Santoni}, \binits{A.}},
\bauthor{\bsnm{Sanz}, \binits{R.B.}},
\bauthor{\bsnm{Saresella}, \binits{M.}},
\bauthor{\bsnm{Sautes-Fridman}, \binits{C.}},
\bauthor{\bsnm{Sawitzki}, \binits{B.}},
\bauthor{\bsnm{Schadt}, \binits{L.}},
\bauthor{\bsnm{Scheffold}, \binits{A.}},
\bauthor{\bsnm{Scherer}, \binits{H.U.}},
\bauthor{\bsnm{Schiemann}, \binits{M.}},
\bauthor{\bsnm{Schildberg}, \binits{F.A.}},
\bauthor{\bsnm{Schimisky}, \binits{E.}},
\bauthor{\bsnm{Schlitzer}, \binits{A.}},
\bauthor{\bsnm{Schlosser}, \binits{J.}},
\bauthor{\bsnm{Schmid}, \binits{S.}},
\bauthor{\bsnm{Schmitt}, \binits{S.}},
\bauthor{\bsnm{Schober}, \binits{K.}},
\bauthor{\bsnm{Schraivogel}, \binits{D.}},
\bauthor{\bsnm{Schuh}, \binits{W.}},
\bauthor{\bsnm{Schüler}, \binits{T.}},
\bauthor{\bsnm{Schulte}, \binits{R.}},
\bauthor{\bsnm{Schulz}, \binits{A.R.}},
\bauthor{\bsnm{Schulz}, \binits{S.R.}},
\bauthor{\bsnm{Scottá}, \binits{C.}},
\bauthor{\bsnm{Scott-Algara}, \binits{D.}},
\bauthor{\bsnm{Sester}, \binits{D.P.}},
\bauthor{\bsnm{Shankey}, \binits{T.V.}},
\bauthor{\bsnm{Silva-Santos}, \binits{B.}},
\bauthor{\bsnm{Simon}, \binits{A.K.}},
\bauthor{\bsnm{Sitnik}, \binits{K.M.}},
\bauthor{\bsnm{Sozzani}, \binits{S.}},
\bauthor{\bsnm{Speiser}, \binits{D.E.}},
\bauthor{\bsnm{Spidlen}, \binits{J.}},
\bauthor{\bsnm{Stahlberg}, \binits{A.}},
\bauthor{\bsnm{Stall}, \binits{A.M.}},
\bauthor{\bsnm{Stanley}, \binits{N.}},
\bauthor{\bsnm{Stark}, \binits{R.}},
\bauthor{\bsnm{Stehle}, \binits{C.}},
\bauthor{\bsnm{Steinmetz}, \binits{T.}},
\bauthor{\bsnm{Stockinger}, \binits{H.}},
\bauthor{\bsnm{Takahama}, \binits{Y.}},
\bauthor{\bsnm{Takeda}, \binits{K.}},
\bauthor{\bsnm{Tan}, \binits{L.}},
\bauthor{\bsnm{Tárnok}, \binits{A.}},
\bauthor{\bsnm{Tiegs}, \binits{G.}},
\bauthor{\bsnm{Toldi}, \binits{G.}},
\bauthor{\bsnm{Tornack}, \binits{J.}},
\bauthor{\bsnm{Traggiai}, \binits{E.}},
\bauthor{\bsnm{Trebak}, \binits{M.}},
\bauthor{\bsnm{Tree}, \binits{T.I.M.}},
\bauthor{\bsnm{Trotter}, \binits{J.}},
\bauthor{\bsnm{Trowsdale}, \binits{J.}},
\bauthor{\bsnm{Tsoumakidou}, \binits{M.}},
\bauthor{\bsnm{Ulrich}, \binits{H.}},
\bauthor{\bsnm{Urbanczyk}, \binits{S.}},
\bauthor{\bparticle{van~de} \bsnm{Veen}, \binits{W.}},
\bauthor{\bparticle{van~den} \bsnm{Broek}, \binits{M.}},
\bauthor{\bparticle{van~der} \bsnm{Pol}, \binits{E.}},
\bauthor{\bsnm{Van~Gassen}, \binits{S.}},
\bauthor{\bsnm{Van~Isterdael}, \binits{G.}},
\bauthor{\bparticle{van} \bsnm{Lier}, \binits{R.A.W.}},
\bauthor{\bsnm{Veldhoen}, \binits{M.}},
\bauthor{\bsnm{Vento-Asturias}, \binits{S.}},
\bauthor{\bsnm{Vieira}, \binits{P.}},
\bauthor{\bsnm{Voehringer}, \binits{D.}},
\bauthor{\bsnm{Volk}, \binits{H.-D.}},
\bauthor{\bparticle{von} \bsnm{Borstel}, \binits{A.}},
\bauthor{\bparticle{von} \bsnm{Volkmann}, \binits{K.}},
\bauthor{\bsnm{Waisman}, \binits{A.}},
\bauthor{\bsnm{Walker}, \binits{R.V.}},
\bauthor{\bsnm{Wallace}, \binits{P.K.}},
\bauthor{\bsnm{Wang}, \binits{S.A.}},
\bauthor{\bsnm{Wang}, \binits{X.M.}},
\bauthor{\bsnm{Ward}, \binits{M.D.}},
\bauthor{\bsnm{Ward-Hartstonge}, \binits{K.A.}},
\bauthor{\bsnm{Warnatz}, \binits{K.}},
\bauthor{\bsnm{Warnes}, \binits{G.}},
\bauthor{\bsnm{Warth}, \binits{S.}},
\bauthor{\bsnm{Waskow}, \binits{C.}},
\bauthor{\bsnm{Watson}, \binits{J.V.}},
\bauthor{\bsnm{Watzl}, \binits{C.}},
\bauthor{\bsnm{Wegener}, \binits{L.}},
\bauthor{\bsnm{Weisenburger}, \binits{T.}},
\bauthor{\bsnm{Wiedemann}, \binits{A.}},
\bauthor{\bsnm{Wienands}, \binits{J.}},
\bauthor{\bsnm{Wilharm}, \binits{A.}},
\bauthor{\bsnm{Wilkinson}, \binits{R.J.}},
\bauthor{\bsnm{Willimsky}, \binits{G.}},
\bauthor{\bsnm{Wing}, \binits{J.B.}},
\bauthor{\bsnm{Winkelmann}, \binits{R.}},
\bauthor{\bsnm{Winkler}, \binits{T.H.}},
\bauthor{\bsnm{Wirz}, \binits{O.F.}},
\bauthor{\bsnm{Wong}, \binits{A.}},
\bauthor{\bsnm{Wurst}, \binits{P.}},
\bauthor{\bsnm{Yang}, \binits{J.H.M.}},
\bauthor{\bsnm{Yang}, \binits{J.}},
\bauthor{\bsnm{Yazdanbakhsh}, \binits{M.}},
\bauthor{\bsnm{Yu}, \binits{L.}},
\bauthor{\bsnm{Yue}, \binits{A.}},
\bauthor{\bsnm{Zhang}, \binits{H.}},
\bauthor{\bsnm{Zhao}, \binits{Y.}},
\bauthor{\bsnm{Ziegler}, \binits{S.M.}},
\bauthor{\bsnm{Zielinski}, \binits{C.}},
\bauthor{\bsnm{Zimmermann}, \binits{J.}},
\bauthor{\bsnm{Zychlinsky}, \binits{A.}}:
\batitle{Guidelines for the use of flow cytometry and cell sorting in
  immunological studies (second edition)}.
\bjtitle{European Journal of Immunology}
\bvolume{49}(\bissue{10}),
\bfpage{1457}--\blpage{1973}
(\byear{2019})
{\href{https://arxiv.org/abs/https://onlinelibrary.wiley.com/doi/pdf/10.1002/eji.201970107}{{https://onlinelibrary.wiley.com/doi/pdf/10.1002/eji.201970107}}}.
\doiurl{10.1002/eji.201970107}
\end{barticle}
\endbibitem

\end{thebibliography}


%%\input bib.tex%

\end{document}